%% file: main.tex
\documentclass[10pt,twocolumn,letterpaper]{article}

\usepackage[algorithms]{wacv}      

\usepackage{graphicx}
\usepackage{amsmath,amssymb}
\usepackage{booktabs}
\usepackage[none]{hyphenat}
\usepackage{array}
\usepackage{multirow}
\usepackage{makecell}
\usepackage{colortbl}
\usepackage{threeparttable}
\usepackage{xurl}           
\usepackage{times}          
\usepackage[normalem]{ulem} 
\usepackage{natbib}
\usepackage[dvipsnames]{xcolor}
\definecolor{wacvblue}{rgb}{0.21,0.49,0.74}

\usepackage{pgfplots}
\usepackage{pgfplotstable}
\pgfplotsset{compat=1.17}

\usepackage{algorithm}
\usepackage{algpseudocode}

\usepackage{caption}
\newcommand{\cmark}{\textcolor{green}{\checkmark}}
\newcommand{\xmark}{\textcolor{red}{\texttimes}}

\input{preamble}

\usepackage[pagebackref,breaklinks,colorlinks,allcolors=wacvblue]{hyperref}

\usepackage[capitalize]{cleveref}

\crefname{section}{Sec.}{Secs.}
\Crefname{section}{Section}{Sections}
\crefname{table}{Tab.}{Tabs.}
\Crefname{table}{Table}{Tables}

\def\wacvPaperID{301} 
\def\confName{WACV}
\def\confYear{2026}
\begin{document}

\title{TalkingPose: Efficient Face and Gesture Animation \\ with Feedback-guided Diffusion Model}

\author{
Alireza Javanmardi$^{1}$ \quad
Pragati Jaiswal$^{1,2}$ \quad
Tewodros Amberbir Habtegebrial$^{1,2}$ \quad
Christen Millerdurai$^{1}$ \\
Shaoxiang Wang$^{1,2}$ \quad
Alain Pagani$^{1}$ \quad
Didier Stricker$^{1,2}$\\[4pt]
$^{1}$German Research Center for Artificial Intelligence (DFKI) 
$^{2}$RPTU\\[4pt]
}

\twocolumn[{%
\renewcommand\twocolumn[1][]{#1}%
\maketitle

\vspace{-25pt}
\begin{center}
    \centering
    \captionsetup{type=figure}
    \includegraphics[width=1\textwidth]{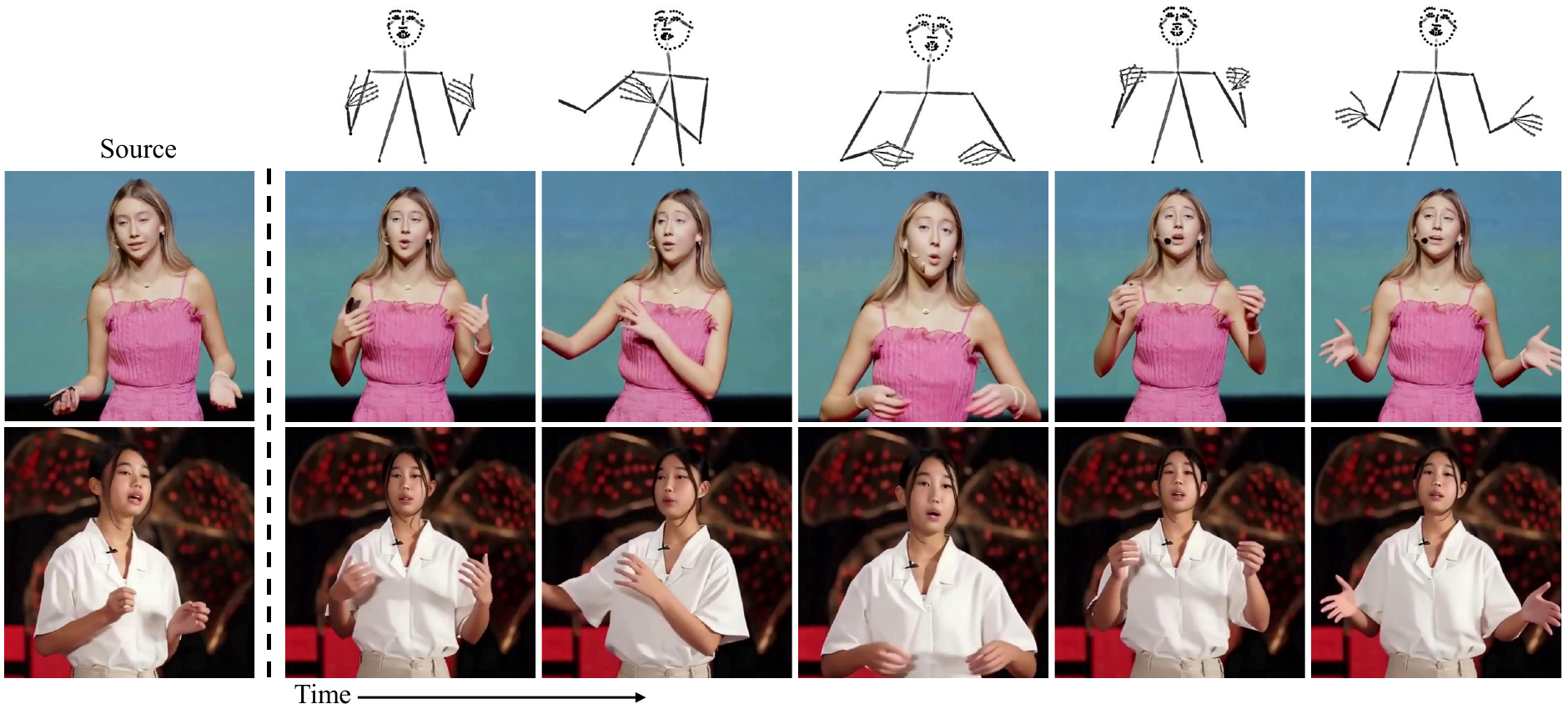}
    \vspace{-19pt} 
    \captionof{figure}{\textbf{TalkingPose} provides limitless face and gesture animation, while achieving the highest efficiency among current video diffusion approaches. Here, we animate the source characters (single frame input) based on the driving motions.}
\end{center}%
}]

\maketitle
\input{sec/00_abstract.tex}

\input{sec/01_intro.tex}
\input{sec/02_related_work.tex}

\input{sec/03_method.tex}

\input{sec/04_experiments.tex}

\input{sec/05_conclusion.tex}
{
    \small
    \bibliographystyle{ieeenat_fullname}
    \bibliography{main}
}
\clearpage
\appendix
\input{sec/X_suppl}

\end{document}

%% file: preamble.tex
\usepackage{xcolor}
\usepackage{soul}
\definecolor{darkyellow}{RGB}{204,153,0} 

\sethlcolor{yellow!40} 

\newcommand{\AJh}[1]{\textcolor{black}{#1}}

%% file: sec/00_abstract.tex
\begin{abstract}
Recent advancements in diffusion models have significantly improved the realism and generalizability of character-driven animation, enabling the synthesis of high-quality motion from just a single RGB image and a set of driving poses. Nevertheless, generating temporally coherent long-form content remains challenging. Existing approaches are constrained by computational and memory limitations, as they are typically trained on short video segments, thus performing effectively only over limited frame lengths and hindering their potential for extended coherent generation.
To address these constraints, we propose TalkingPose, a novel diffusion-based framework specifically designed for producing long-form, temporally consistent human upper-body animations. TalkingPose leverages driving frames to precisely capture expressive facial and hand movements, transferring these seamlessly to a target actor through a stable diffusion backbone. To ensure continuous motion and enhance temporal coherence, we introduce a feedback-driven mechanism built upon image-based diffusion models. Notably, this mechanism does not incur additional computational costs or require secondary training stages, enabling the generation of animations with unlimited duration. Additionally, we introduce a comprehensive, large-scale dataset to serve as a new benchmark for human upper-body animation. Project page: \url{https://dfki-av.github.io/TalkingPose}
\end{abstract}

%% file: sec/01_intro.tex
\vspace{-15pt}
\section{Introduction}
Animating humans from a single image stems from the increasing demand for authentic, engaging digital interactions across various fields such as virtual reality, gaming, remote communication, and content creation.
This technology has promising applications in virtual communication and the medical industry, especially where hand gestures are essential, such as in sign language understanding or hand-intensive fields like surgery, where precise gestures and facial expressions are crucial.
The core of this challenge involves extracting the appearance of a "target" human actor from a single source image and animating it by transferring motion from a sequence of driving frames. 
Consequently, recent research has focused on developing methods that either animate only the target actor's facial expressions~\cite{wang2021one,ma2024follow, kirschstein2024diffusionavatars} or generate holistic animations of the entire body \cite{siarohin2021motion, wang2022latent}. 
%
Despite significant progress in this domain, two primary obstacles persist. First, the limited availability of large-scale datasets focusing on human upper-body motion constrains the robustness and generalizability of current methods. Second, both training and deploying video generative models remain computationally demanding, thereby restricting their outputs to short frame sequences and inhibiting the seamless synthesis of continuous, long-form animations.

Early methods primarily employed computer graphics techniques targeting the facial region, utilizing parametric face models to animate reconstructed heads \cite{Thies_2016_CVPR, thies2015real}. 
These approaches have seen notable advancements with the introduction of differentiable rendering techniques \cite{liu2019soft}, which blend machine learning with traditional graphics pipelines \cite{prinzler2023diner, zielonka2023instant, qian2024gaussianavatars}.
However, these methods typically relied on multi-view data or RGB-D sensors for accurate face reconstruction. 
With the advancement of generative models, particularly Generative Adversarial Networks (GANs) \cite{goodfellow2014generative}, researchers have applied them to \textit{one-shot} talking head synthesis. 
They typically use keypoints and feature warping techniques, which show impressive results when trained on facial data \cite{siarohin2019first,wang2021one}. 
However, extending these models to upper-body or full-body videos poses challenges in capturing detailed facial expressions and precise hand gestures.

More recently, Diffusion models \cite{ho2020denoising} have demonstrated remarkable capabilities in face and body retargeting, leveraging advanced conditioning information to enable high-resolution character animation with enhanced photorealism \cite{ma2024follow, xu2024magicanimate, hu2024animate}. 
Nevertheless, they face two primary challenges: ensuring consistency across animation sequences (on the order of hundreds of frames) and managing the substantial computational and memory demands of video synthesis.
To address these limitations, we introduce \textit{TalkingPose}, a novel framework that integrates hand gestures into facial animation models via a feedback-guided diffusion backbone. Unlike previous works that rely on temporal layers for consistency \cite{hu2024animate,chang2023magicpose,xu2024magicanimate,zhu2024champ}—requiring either training on stacks of frames \cite{mimicmotion2024,tu2024stableanimator} or, training single-frame models followed by fine-tuning only the temporal layers \cite{hu2024animate,zhu2024champ}—both approaches demand substantial computational resources and large-scale, temporally consistent video input.
In contrast, we propose a closed-loop control (CLC) mechanism applied solely during inference, which significantly reduce frame-to-frame inconsistencies without the need for additional training or parameters. This design notably reduces computational overhead while enabling continuous, long-form animation. Furthermore, we introduce a large-scale video dataset encompassing upper-body motions—including diverse facial expressions, hand gestures, appearances, and backgrounds \\
In summary, our key contributions are as follows: 
\begin{itemize} \item We propose a closed-loop control mechanism that stabilizes diffusion models during inference, ensuring robust temporal consistency for extended video generation while significantly improving computational efficiency.
 \item We present \textit{TalkingPose}, a large-scale video dataset covering diverse ages, genders, facial expressions, hand gestures, appearances, and backgrounds. \item We evaluate our method on the TED-talk \cite{siarohin2021motion}, TikTok \cite{jafarian2021learning} datasets and our newly collected \textit{TalkingPose}, showing its effectiveness through both qualitative and quantitative results. \end{itemize}

%% file: sec/02_related_work.tex
\section{Related Works}

\subsection{Diffusion Models for Video Generation}
Diffusion models gained significant attention with Denoising Diffusion Probabilistic Models (DDPMs) \cite{ho2020denoising} as powerful generative models for image generation tasks \cite{dhariwal2021diffusion, ho2022cascaded}. 
Latent diffusion models \cite{rombach2022high} later improved efficiency, paving the way for applications in video generation \cite{esser2023structure, blattmann2023stable, guo2023animatediff}.
These models typically use U-Net \cite{ronneberger2015u} or transformer architectures \cite{vaswani2017attention} to perform diffusion on stacks of latent embeddings. 
For conditional video generation, methods often employ CLIP \cite{radford2021learning} for text/image conditioning, with ControlNet \cite{zhang2023adding} and T2I-Adapter \cite{mou2024t2i} enhancing control over attributes such as pose, mask, and edge. \\
Early works attempted to generate consistent videos by training on sequences of frames using 3D U-Net models \cite{ho2022video}. 
Later works introduced an additional temporal attention layer \cite{guo2023animatediff}, which performs attention along the temporal dimension. 
While these strategies improve temporal coherence, they also demand extensive temporal data training, which can be computationally and memory-intensive \cite{hu2024animate, chang2023magicpose, xu2024magicanimate}. Moreover, at inference, they often struggle to preserve consistency across successive frames, especially for longer videos. To address this, several works generate longer videos in a coarse-to-fine manner \cite{yin2023nuwa, wang2024magicvideo} or progressively merge latent features of overlapping frames of video chunks \cite{mimicmotion2024}, thereby enhancing transitions between frames. Nevertheless, these methods are predominantly designed to smooth transitional segments rather than comprehensively handle extended video generation. To overcome these limitations, we introduce a feedback-driven mechanism that enables consistent, continuous character video generation without extra computational overhead.

\subsection{Face and Body Animation Generation}
The field of face and body animation began with early approaches that primarily relied on graphics-based techniques \cite{Thies_2016_CVPR, thies2015real}, such as parametric face models like FLAME \cite{FLAME:SiggraphAsia2017} for facial animation and SMPL \cite{SMPL:2015} for full-body reconstruction. Although these models enabled animatable 3D avatars, they often struggled with finer details, such as hair and accessories (e.g., glasses). Recent approaches based on implicit neural representations \cite{sitzmann2020implicit, mildenhall2021nerf} and 3D Gaussian Splatting (3DGS) \cite{kerbl20233d} have greatly enhanced photorealism and rendering speed \cite{prinzler2023diner, zielonka2023instant, xu2024gaussian, qian2024gaussianavatars}. However, these methods typically require multi-view training data and substantial training time for each individual avatar, limiting their scalability in broader applications. \\ 
In parallel, researchers have explored generative models that learn the underlying distribution of avatars from large-scale video datasets, enabling single-shot, instantaneous inference \cite{wang2021one,siarohin2019first,yang2022face2face}. \\
Early efforts in this domain focused on feature-warping techniques using facial keypoints in an adversarial manner \cite{siarohin2019first,wang2021one} While these self-supervised models perform well in talking head synthesis, they show limitations when trained on upper-body or full-body video datasets \cite{siarohin2019first}. Some works have improved the framework by incorporating body regions \cite{siarohin2021motion} or inverse GAN techniques \cite{wang2022latent}, yet they show limitations in capturing intricate details, particularly in facial expressions and hand movements simultaneously. \\
Recently, diffusion-based models have become promising alternatives for animating human faces. While they rely on driving frames, these methods also integrate other modalities such as audio \cite{xu2024vasa, lin2024cyberhost,jiang2024loopy,tian2024emo}, and have shown strong results in generating dancing avatars guided by music or motion sequences \cite{kim2024tcan,mimicmotion2024,xu2024magicanimate,zhu2024champ}. A separate line uses diffusion to map a driving actor’s appearance onto a source video, as in MIMO and AnimateAnyone2 \cite{men2025mimo, hu2025animateanyone2}.
For appearance extraction, many recent methods adopt an “Appearance/Reference Net” \cite{zhu2024champ,chang2023magicpose,hu2024animate}, which replicates the Stable Diffusion U-Net and processes the source image in parallel. They then either copy or concatenate the spatial attention layers from the Appearance Net to the main U-Net—sometimes by themselves \cite{chang2023magicpose,xu2024magicanimate} or in combination with CLIP and cross-attention modules \cite{hu2024animate}. Other approaches refine specific aspects of the avatar by employing SMPL \cite{SMPL:2015} to improve shape and pose, or by incorporating a face mask region and ArcFace features \cite{tu2024stableanimator} to enhance facial realism.

%% file: sec/03_method.tex
\section{Method}
In our TalkingPose approach, we animate a target actor from a given source image using a sequence of driving frames that capture the desired motion. 
We begin with an overview of Stable Diffusion in Sec.\ref{subsec:preliminaries}, followed by a description of our framework in Sec.\ref{subsec:framework}, an explanation of our feedback mechanism during inference in Sec.\ref{subsec:inference}, and an introduction to our dataset in Sec.\ref{subsec:talking_pose_dataset}.
\begin{figure*}[t]
  \centering
  \vspace{-15pt}
  \includegraphics[width=2.1\columnwidth]{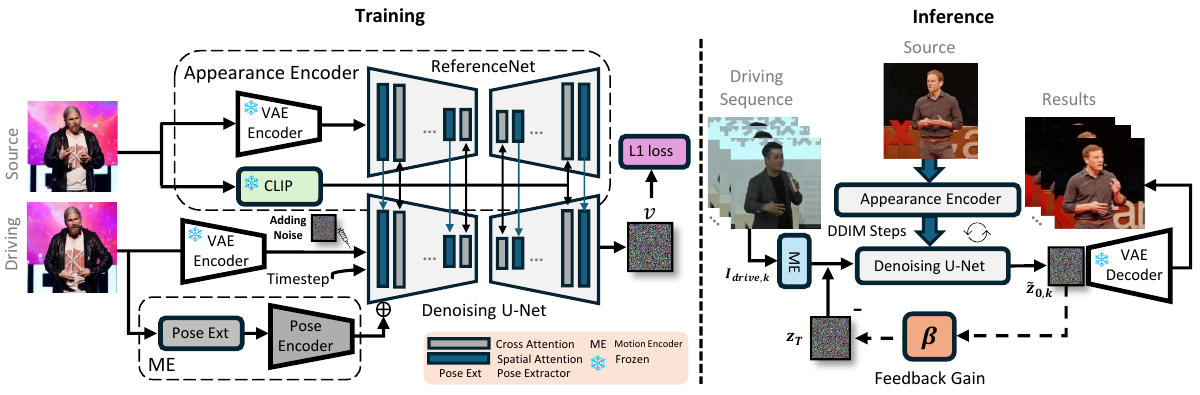}
  \captionsetup{skip=5pt} 
  \vspace{-15pt}
\caption{
\textbf{TalkingPose Pipeline.} \textbf{Training:} Following AnimateAnyone \cite{hu2024animate}, the Appearance Encoder (CLIP + ReferenceNet) obtains source features, while the Motion Encoder (driving pose extraction using method of Yang et al.~\cite{yang2023effective} + Pose Encoder) prepares motion cues for the U-net. \textbf{Inference:} A single RGB source and a driving pose condition drive DDIM steps to predict a latent, which is refined via a feedback loop with proportional gain (\(\beta\)).
}
\vspace{-15pt}
  \label{fig:pipeline}
\end{figure*}

\subsection{Preliminaries}
\label{subsec:preliminaries}
Stable Diffusion~\cite{rombach2022high} operates on a learned latent space, substantially reducing computational costs versus pixel-space diffusion. 
An input image \(I \in \mathbb{R}^{H \times W \times C}\) is encoded into a latent representation \(z = E(I) \in \mathbb{R}^{h \times w \times c}\) by a VAE encoder \(E\)~\cite{kingma2014auto}, with a downsampling factor \(f = \tfrac{H}{h} = \tfrac{W}{w} = 2^m,\, m \in \mathbb{N}\).
The forward diffusion process incrementally adds Gaussian noise, \(\epsilon \sim \mathcal{N}(0,I)\), at each step \(t\) (drawn uniformly from \(\{1, \dots, T\}\)), producing a noisy latent \(z_t\).
A U-Net-based denoising network \(\epsilon_\theta\) then estimates \(\epsilon\) from \(\bigl(z_t, t, \tau_\theta(y)\bigr)\) or uses \emph{v-prediction} to estimate a weighted combination of the noise and the clean latent \(x_0\). Here, \(\tau_\theta\) is a domain-specific encoder (e.g., CLIP~\cite{radford2021learning}) that maps the image \(y\) to a conditioning vector.
During training, the objective
\[
L_{\text{denoise}} 
= \mathbb{E}_{z,\;\epsilon,\;t,\;y}
\!\Bigl[
\|\epsilon \;-\; \epsilon_\theta\bigl(z_t,\,t,\,\tau_\theta(y)\bigr)\|^2
\Bigr]
\]
enforces accurate noise prediction, enabling effective inversion of the forward process. During inference, denoising steps (e.g., DDIM~\cite{song2020denoising}) generate~$\tilde{z}_0 \approx z$, which is decoded by~$D$ to reconstruct the final image~$\tilde{x}_0 = D(\tilde{z}_0)$. This approach enables high-fidelity image generation with reduced computational overhead.

\subsection{TalkingPose Framework}
\label{subsec:framework}
As Illustrated in Fig.~\ref{fig:pipeline}, our method takes as input 
\begin{enumerate*}[label=(\arabic*)]
  \item a source image showing the target actor’s appearance, and 
  \item a sequence of driving frames that convey the desired motion.
\end{enumerate*}
The objective is to generate new frames of the target actor, preserving the subject’s appearance while adopting the motion from the driving frames.
Building upon the AnimateAnyone pipeline~\cite{hu2024animate}, we remove the temporal attention layers to enable training on individual frames rather than complete video sequences. 
In each training iteration, two frames are randomly selected from a video: the first captures the actor's appearance as the source frame, while the second provides the motion cues as the driving frame.
Both frames are converted into latent representations via a VAE encoder. 
The source latent is input to the ReferenceNet~\cite{hu2024animate}.
Additionally, the source frame is processed by a CLIP ViT-L/14 encoder~\cite{radford2021learning} to extract appearance features, which are injected into both the Denoising U-Net and the ReferenceNet via cross-attention mechanism. We refer to this integrated module as the \emph{Appearance Encoder}.
Simultaneously, the motion encoder extracts motion features from the driving frame.
To simulate the forward diffusion process, noise is added to the latent representation of the driving frame based on the selected time-step. 
This noisy latent, along with the motion features, is then input to the Denoising U-Net. 
The network is then trained to predict the added noise, with the discrepancy supervised by an $L_1$  loss. 
By accurately recovering the noise, the U-Net effectively fuses the source appearance with the driving motion. For more architectural details, readers are referred to AnimateAnyone~\cite{hu2024animate}.

\subsection{Inference with Closed-loop Control}
\label{subsec:inference}
Generating long-form content with diffusion models poses significant computational challenges, particularly in terms of GPU resource utilization during both the training and inference phases. The memory requirements increase proportionally with the number of frames being processed.
Additionally, segmenting videos into smaller parts can lead to cumulative error propagation, since using only the initial or final frame of a segment as the “source” may amplify discrepancies over time. 
Furthermore, the batch-wise approach tends to introduce temporal artifacts during transitions between batches.

To address these limitations, we introduce \emph{closed-loop control} (CLC), a novel feedback-driven inference strategy that can be seamlessly integrated with any latent diffusion-based model. 
As illustrated in Fig.~\ref{fig:pipeline}, we begin by extracting the appearance from a source image---potentially distinct from the identity depicted in the driving pose---and merging it with the target pose from the first driving frame.
Specifically, we generate a motion encoding for this target pose and add it to Gaussian noise to create the input for our U-Net. 
Using DDIM steps, we obtain an output latent encoding that fuses the source image’s appearance with the motion from the first driving frame. 
This latent is then passed through the Stable Diffusion VAE decoder to generate the initial frame of the animation.
For subsequent frames, we reintroduce the generated output latent encoding into the sampled noise, regulated by a feedback gain to maintain a consistent appearance of the avatar across frames.
This feedback gain is modeled from a control theory perspective, treating it as a closed-loop control system~\cite{kailath1980linear, xu2020understanding},
Specifically, we use negative feedback to reduce unexpected deviations, counteracting small errors, minimizing oscillations, and guiding the output toward a stable, consistent appearance.
The process works as follows: for each frame $k$, the diffusion model requires a Gaussian noise $z_T$  as the initial input for inference. 
The diffusion process runs over several timesteps to produce the output latent encoding $\hat{z}_{0,k}$ for the frame $k$. 
To generate the next frame, $\hat{z}_{0,k+1}$, we apply the following state update equation:
\begin{equation}  
x_{k+1} = z_T + \beta ( \hat{z}_{0,k} - z_T )
\label{eq:eq7}
\end{equation}
where $x_{k+1}$ is the U-net input before adding to the motion encoding to produce $\hat{z}_{0,k+1}$, and $\beta$ is the feedback gain parameter.
This feedback loop reduces errors and disturbances, ensuring consistency across generated frames over time.
This approach is significantly more efficient than methods relying on temporal attention~\cite{hu2024animate, chang2023magicpose, xu2024magicanimate}, which can also suffer from inconsistencies over extended periods. 
Additionally, this inference-based method eliminates the need for extra training and additional parameters while enabling the generation of an unlimited number of stable, consistent frames. Further details of our method is provided in \cref{alg:clc}.

\begin{algorithm}[h]
\caption{Inference with Closed-loop Control (CLC)}
\label{alg:clc}
\begin{algorithmic}
\Require
  \Statex $I_{\text{source}}$: source image
  \Statex $\{I_{\text{drive},k}\}_{k=1}^{\infty}$: driving images
  \Statex $T$: number of diffusion steps
  \Statex $\beta$: feedback gain
\State $a_{\text{src}} \leftarrow \text{Appearance Encoder}(I_{\text{source}})$
\State \textbf{Sample} $z_T \sim \mathcal{N}(0,I)$
\State $x_{1} \leftarrow z_T$
\For{$k = 1$ to $\infty$}
  \State $m_k \leftarrow \text{MotionEncoder}(I_{\text{drive},k})$
  \For{$t = T$ down to $1$}
    \State $\hat{z}_{t,k} \leftarrow DM_\theta\bigl(x_k, a_{\text{src}}, m_k; t\bigr)$
  \EndFor
  \State $x_{k+1} \leftarrow z_T + \beta\,\bigl(\hat{z}_{0,k} - z_T\bigr)$
  \State $\hat{I}_k \leftarrow \text{Decoder}\bigl(\hat{z}_{0,k}\bigr)$
\EndFor
\end{algorithmic}
\end{algorithm}
\subsection{TalkingPose Dataset} \label{subsec:talking_pose_dataset}
We curate a large-scale dataset of human upper-body videos from diverse YouTube sources, aiming to capture not only facial geometry and expressions but also hand articulation, body movements, clothing diversity, and varying backgrounds. This scope goes beyond face-focused datasets, where only facial attributes are emphasized. As a result, learning generative models from upper-body data is more challenging, requiring robust representation of a broader set of visual and motion cues.
Our initial collection included 21K raw videos, which, after a multi-stage preprocessing pipeline, yielded \textit{18K videos with unique identities}. Each video typically features a single presenter centered in the frame. From these videos, we detect and crop the upper-body region at a resolution of {512$\times$512}. Samples failing to meet this criterion (e.g., insufficient resolution, missing face or hands) are discarded. The final dataset spans approximately 1250 hours of video footage at 20~FPS, encompassing diverse locations, lighting conditions, ages, and a balanced gender distribution. A detailed breakdown of the preprocessing steps and demographic statistics is provided in the Appendix.

\begin{figure*}[!t]
\vspace{-15pt}
  \centering
  \includegraphics[width=2.05\columnwidth]{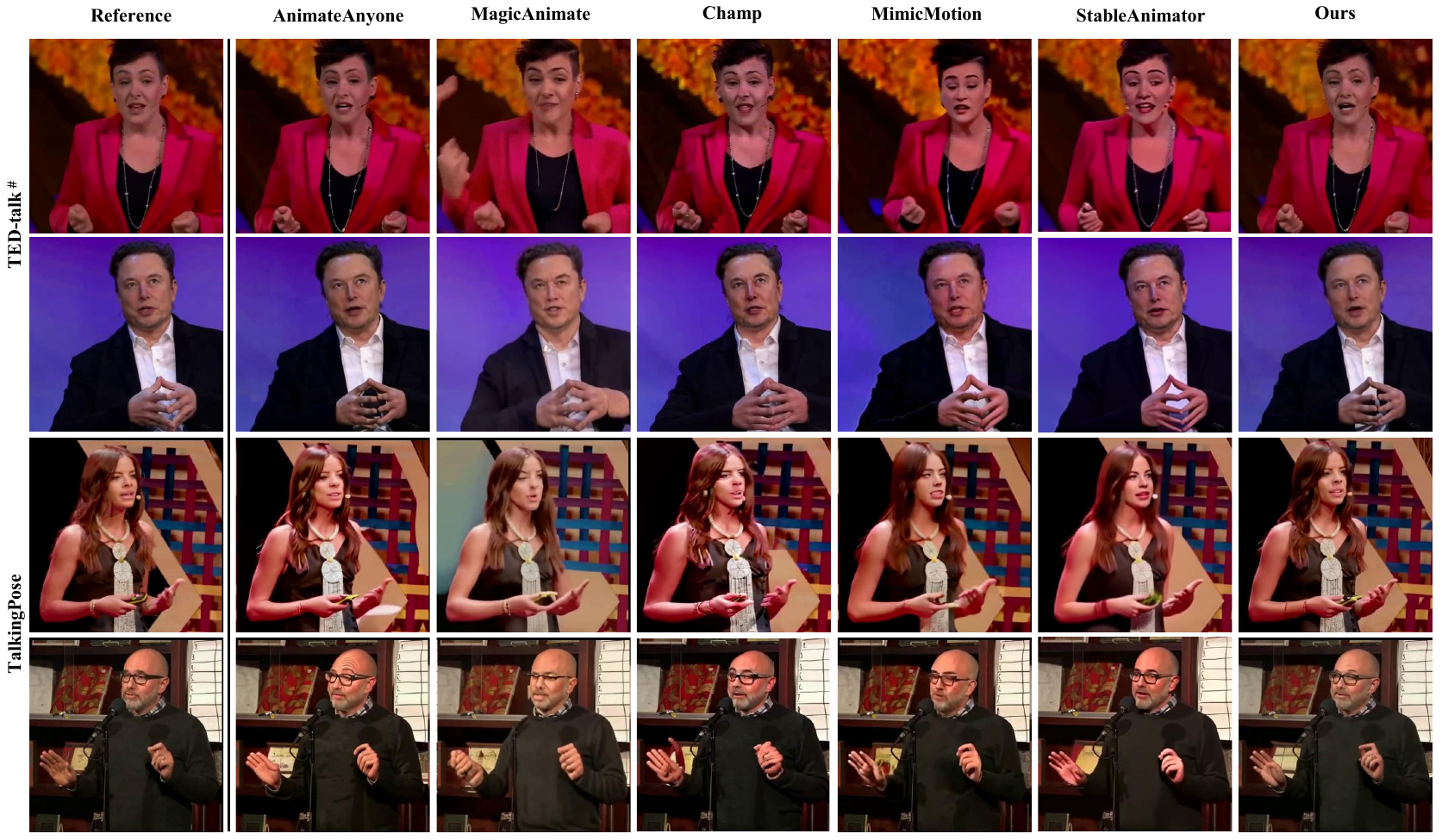}
  \captionsetup{skip=5pt} 
\vspace{-7pt}
\caption{\textbf{Qualitative Comparison}. This figure shows our model's ability to accurately capture facial expressions, hand gestures, and poses across diverse postures, while preserving the appearance and background of the reference frame, compared to state-of-the-art methods AnimateAnyone \cite{hu2024animate}, MagicAnimate \cite{xu2024magicanimate}, \AJh{Champ} \cite{zhu2024champ}, MimicMotion \cite{mimicmotion2024}, and StableAnimator \cite{tu2024stableanimator} on TED-talk\textsuperscript{\#} \cite{siarohin2021motion} and TalkingPose.}
  \label{fig: qualitative}
\end{figure*}

\begin{table*}[!t]
\centering
\scriptsize
\setlength{\tabcolsep}{5.2pt}
\renewcommand{\arraystretch}{1.08}
\resizebox{\textwidth}{!}{%
\begin{tabular}{l ccccc c cccc}
\toprule
\multirow{2}{*}{\textbf{Method}}
& \multicolumn{5}{c}{\textbf{\AJh{TED-talk\textsuperscript{\#}}}}
& \multicolumn{5}{c}{\textbf{TalkingPose}} \\
& {\scriptsize \textbf{SSIM $\uparrow$}}
& {\scriptsize \textbf{\AJh{PSNR\textsubscript{float}} $\uparrow$}}
& {\scriptsize \textbf{LPIPS $\downarrow$}}
& {\scriptsize \textbf{FID-VID $\downarrow$}}
& {\scriptsize \textbf{FVD $\downarrow$}}
& {\scriptsize \textbf{SSIM $\uparrow$}}
& {\scriptsize \textbf{PSNR\textsubscript{float} $\uparrow$}}
& {\scriptsize \textbf{LPIPS $\downarrow$}}
& {\scriptsize \textbf{FID-VID $\downarrow$}}
& {\scriptsize \textbf{FVD $\downarrow$}} \\
\midrule
AnimateAnyone (Moore impl.)$^{\dagger}$~\cite{hu2024animate}
& 0.768 & \textbf{22.22} & \textbf{0.226} & \underline{5.99} & \textbf{89.90}
& \underline{0.736} & \underline{21.87} & \underline{0.197} & \underline{9.56} & \textbf{192.78} \\
\textcolor{gray}{AnimateAnyone (TED-talk)}
& \textcolor{gray}{0.832} & \textcolor{gray}{--} & \textcolor{gray}{0.159} & \textcolor{gray}{--} & \textcolor{gray}{80.50}
& & & & & \\
MagicPose~\cite{chang2023magicpose}
& 0.638 & 16.25 & 0.307 & 15.16 & 329.44
& 0.509 & 14.27 & 0.305 & 23.57 & 348.05 \\
MagicAnimate~\cite{xu2024magicanimate}
& 0.574 & 15.02 & 0.326 & 19.05 & 401.60
& 0.445 & 12.59 & 0.339 & 31.70 & 425.87 \\
MimicMotion~\cite{mimicmotion2024}
& 0.643 & 16.21 & 0.349 & 12.94 & 297.28
& 0.640 & 17.53 & 0.275 & 12.46 & 274.16 \\
Champ~\cite{zhu2024champ}
& 0.685 & 18.63 & 0.290 & 12.46 & 217.87
& 0.654 & 18.66 & 0.294 & 15.14 & 304.79 \\
StableAnimator~\cite{tu2024stableanimator}
& 0.703 & 19.56 & 0.279 & 12.22 & 203.90
& 0.669 & 18.63 & 0.248 & 16.52 & 276.77 \\
\midrule
TalkingPose (Ours)
& \textbf{0.773} & \underline{21.92} & \underline{0.230} & \textbf{5.32} & \underline{105.62}
& \textbf{0.749} & \textbf{21.94} & \textbf{0.186} & \textbf{5.07} & \underline{203.11} \\
\bottomrule
\end{tabular}%
}
\vspace{-4pt}
\caption{\textbf{Quantitative results on TED-talk\textsuperscript{\#}
\cite{siarohin2021motion} and TalkingPose, evaluated with Disco \cite{wang2024disco}}. 
Best results are in \textbf{bold}, second-best are \underline{underlined}. 
\AJh{For AnimateAnyone, we report both results from the original publication (gray, TED-talk) and our reproduction using the unofficial Moore implementation on TED-talk\textsuperscript{\#}, a variant of TED-talk where missing videos were supplemented.}}
\label{Table:comparison_combined}
\vspace{-8pt}
\end{table*}

\begin{figure*}[h]
\vspace{-15pt}
  \centering
  \includegraphics[width=2\columnwidth]{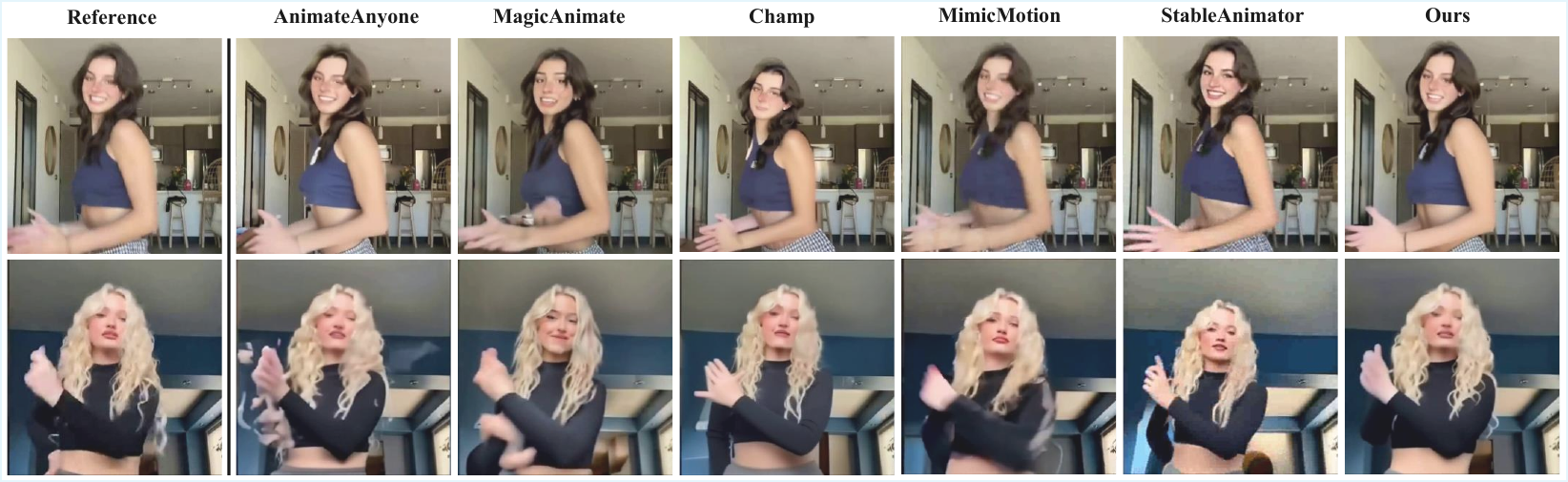}
  \captionsetup{skip=5pt} 
\vspace{-4pt}
\caption{\textbf{Qualitative Comparison}. This figure demonstrates our model's performance compared to state-of-the-art methods AnimateAnyone \cite{hu2024animate}, MagicAnimate \cite{xu2024magicanimate}, \AJh{Champ} \cite{zhu2024champ}, MimicMotion \cite{mimicmotion2024} and StableAnimator \cite{tu2024stableanimator} on TikTok dataset \cite{jafarian2021learning}.}
\vspace{-12pt}
  \label{fig: qualitative2}
\end{figure*}

\begin{table}[!t]
\centering
\small
\setlength{\tabcolsep}{3pt}
\resizebox{0.99\linewidth}{!}{%
\begin{tabular}{@{}lrrrrrrr@{}}
\toprule
\textbf{Method}
 & \multicolumn{1}{c}{\scriptsize SSIM $\uparrow$}
 & \multicolumn{1}{c}{\scriptsize \AJh{PSNR\textsubscript{int}} \footnotesize$\uparrow$\cite{hore2010image}}
 & \multicolumn{1}{c}{\scriptsize \AJh{PSNR\textsubscript{float}}\footnotesize$\uparrow$\cite{wang2024disco}}
 & \multicolumn{1}{c}{\scriptsize LPIPS $\downarrow$}
 & \multicolumn{1}{c}{\scriptsize FID-VID $\downarrow$}
 & \multicolumn{1}{c}{\scriptsize FVD $\downarrow$}\\
\midrule
AnimateAnyone ~\cite{hu2024animate}      
  & 0.718 &  \AJh{29.56}   &  -   & 0.285 & -      & 171.90 \\
MagicPose~\cite{chang2023magicpose}     
  & 0.752 &  \AJh{29.53}  & -  & 0.292 & 46.30  & - \\
MagicAnimate~\cite{xu2024magicanimate}  
  & 0.714 &  \AJh{-}  &  18.22  & 0.239 & 21.75  & 179.07\\
\AJh{MimicMotion}~\cite{mimicmotion2024}   
  & \AJh{0.795} &  \AJh{-}  & \AJh{20.10}   &  \AJh{-}  & \AJh{\textbf{9.30}} & \AJh{594.00}\\
Champ~\cite{zhu2024champ}         
  & \underline{0.802} & \AJh{29.91} & - & 0.234 & 21.07 & \underline{160.82}\\
StableAnimator~\cite{tu2024stableanimator} 
  & 0.801 & \AJh{\textbf{30.81}} & \underline{20.66} & 0.232 & - & \textbf{140.62}\\
\midrule
TalkingPose (Ours) 
  & \textbf{0.822} & \AJh{\underline{30.50}} & \textbf{21.36} & \textbf{0.222} & \underline{15.04} & 226.63 \\
\bottomrule
\end{tabular}%
}
\vspace{-7pt}
\caption{\textbf{Quantitative results on the TikTok dataset \cite{jafarian2021learning}.} Reported values are taken from the corresponding original publications. The PSNR variants are described in Sec.~4.3.}
\vspace{-15pt}
\label{Table:comparison_TikTok}
\end{table}

%% file: sec/04_experiments.tex
\section{Experiments}
\subsection{Datasets}
We conduct our experiments on three benchmark datasets: \\
\noindent\textbf{TED-talk.}~The TED-talk dataset \cite{siarohin2021motion} consists of 369 training and 42 test videos. Since some videos were no longer accessible, we replaced them with comparable recordings from the same YouTube channel; we refer to this version as TED-talk\textsuperscript{\#}.\\
\noindent\textbf{TikTok.}~The TikTok dataset \cite{jafarian2021learning} consists of 350 single-person dance clips, each lasting 10--15 seconds. We follow the original split, using 340 videos for training and 10 for testing. \\
\noindent\textbf{TalkingPose.}~We introduced the \textit{TalkingPose} dataset with about 18K videos of people giving talks with hand gestures, using 90\% for training and 10\% from non-overlapping identities for testing.
\subsection{Implementation Details}
\label{sec:implementation_details}
We trained our model for 70K steps on four NVIDIA H200 GPUs using a batch size of 32 and a learning rate of \(1\times 10^{-5}\) in v-prediction setting \cite{salimans2022progressive}. The CLIP and VAE encoder-decoder weights remained frozen, while we fine-tuned only the ReferenceNet and the Denoising U-net, both initialized from Stable Diffusion~V1.5. The Pose Guider was initialized with Gaussian weights, except for the final projection layer, which employed zero convolution. During inference, we used DDIM sampling with 30 steps and evaluated the model on 50-frame videos using an NVIDIA RTX3090 GPU. To avoid contrast saturation, we set the CFG value to 3.5.
To tune the feedback gain (\(\beta\)) hyperparameter, we performed a grid search on 100 validation videos from the TalkingPose dataset.
\subsection{Qualitative and Quantitative Comparison} 
\label{subsec:comp}
\begin{figure}[h]
  \centering
  \includegraphics[width=0.95\columnwidth]{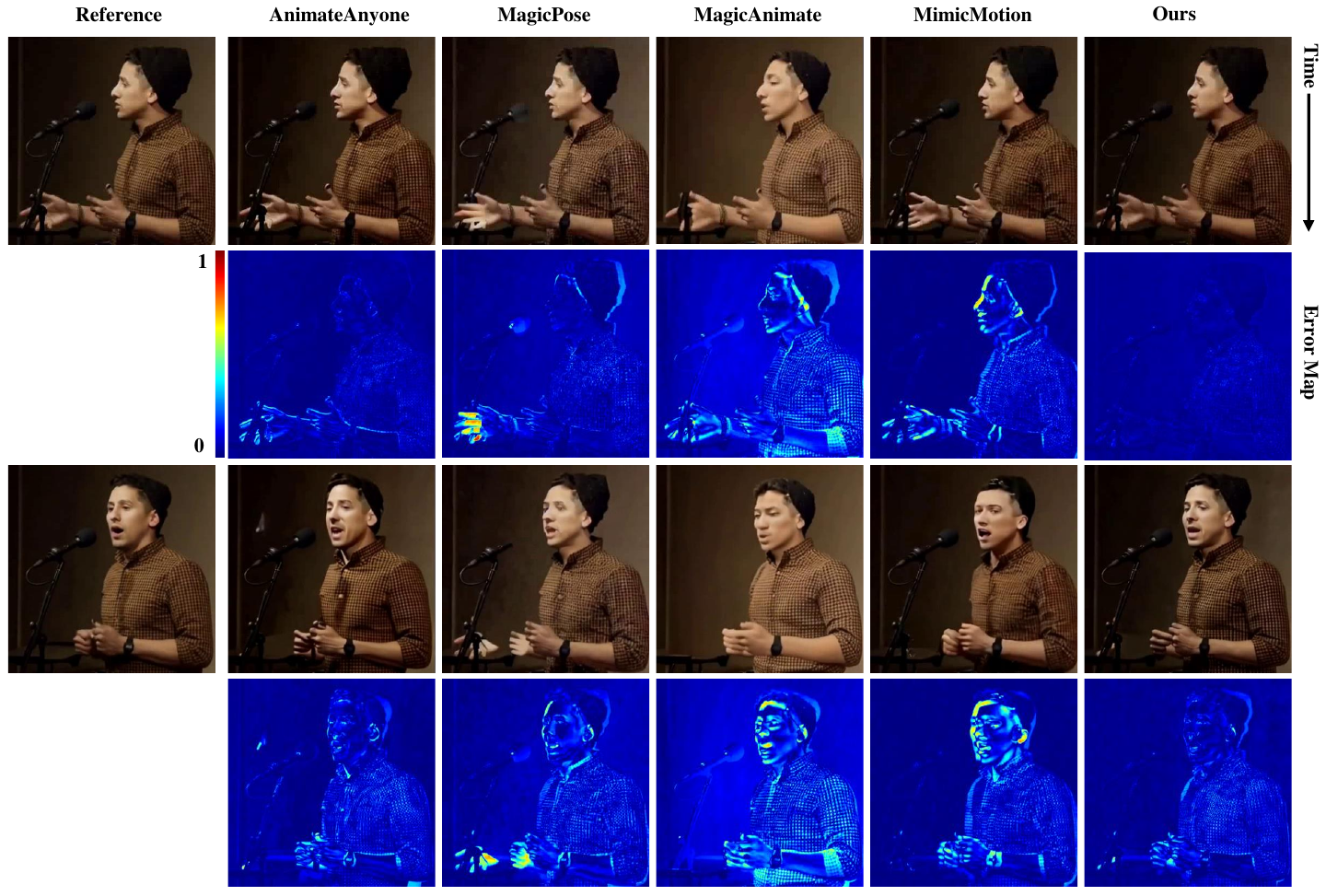}
  \captionsetup{skip=5pt} 
  \vspace{-5pt}
\caption{\textbf{Temporal Qualitative Analysis}. This figure demonstrates temporally consistent character animation and its resemblance to the provided source, along with two randomly selected pose frames from the video. The differences with the reference are highlighted using error map.
}
  \label{fig:temporal_errormap}
\end{figure}

\noindent\textbf{Baseline.} For our comparative analysis, we selected several state-of-the-art models that address both body animation and facial reenactment. Specifically, we evaluated AnimateAnyone \cite{hu2024animate}, MagicPose \cite{chang2023magicpose}, MagicAnimate \cite{xu2024magicanimate}, MimicMotion \cite{mimicmotion2024}, Champ \cite{zhu2024champ} and StableAnimator \cite{tu2024stableanimator}. Since training scripts were unavailable for some methods \cite{xu2024magicanimate,mimicmotion2024}, we relied on official repositories and released pre-trained checkpoints trained on large-scale datasets with an emphasis on generalization for all baselines except AnimateAnyone. For AnimateAnyone, we used the widely recognized unofficial MooreAnimate repository, training both stages.\\ 
\noindent\textbf{Metrics.} We assess image quality using two variants of Peak Signal-to-Noise Ratio (PSNR)~\cite{hore2010image}: \textit{PSNR\textsubscript{int}}, which computes integer-based values as in~\cite{hu2024animate,chang2023magicpose,zhu2024champ}, and \textit{PSNR\textsubscript{float}}, which avoids numerical overflow by using floating-point operations as in~\cite{tu2024stableanimator,mimicmotion2024,xu2024magicanimate}. For our experiments, we mainly report PSNR\textsubscript{float} since it yields more accurate results. We further report Structural Similarity Index Measure (SSIM)~\cite{wang2004image} and Learned Perceptual Image Patch Similarity (LPIPS). To evaluate temporal consistency, we use FID-VID~\cite{balaji2019conditional} and Fréchet Video Distance (FVD)~\cite{unterthiner2019towards}. All metrics are computed using the Disco evaluation toolkit~\cite{wang2024disco}.  
For pose accuracy, we adopt Average Keypoint Distance (AKD)~\cite{JMLR:v12:gashler11a}, using MediaPipe landmarks~\cite{lugaresi2019mediapipe} for face, hands, and torso separately. For facial identity preservation, we compute cosine similarity (CSIM) based on ArcFace embeddings~\cite{deng2019arcface,richardson2021encoding,huang2020curricularface}. Finally, we measure lip synchronization with SyncNet~\cite{chung2016out}.
\\
\noindent\textbf{Evaluation.}
For evaluation, we used the first frame of each video as the source and the full video as the driving sequence, allowing for frame-by-frame comparisons across all models. Methods that rely on temporal layers alleviate short-term flickering, but we observed frequent contrast shifts and frame-specific artifacts (bottom two rows of \cref{fig: qualitative}, top row of \cref{fig: qualitative2} and the supplementary video). We analyse this behaviour in detail in the ablation studies (\cref{subsec:abl}). As shown in \cref{fig: qualitative}, \textit{MagicAnimate} and \textit{MimicMotion} often fail to consistently preserve the character’s facial identity. While other methods reliably preserve facial appearance when generating the video with various expressions.
In terms of pose accuracy, \textit{MagicAnimate} sometimes fails to replicate hand gestures precisely, especially in the TikTok dataset, whereas other methods maintain better pose consistency (see \cref{fig: qualitative2}). Similar issues arise in the TED-talk\textsuperscript{\#} and TalkingPose datasets, where \textit{TalkingPose} again outperforms the competing approaches.
\begingroup
  \renewcommand\thefootnote{}%
  \footnotetext{$^{\dagger}$AnimateAnyone (Moore impl.): \url{https://github.com/MooreThreads/Moore-AnimateAnyone}}%
\endgroup
Quantitatively, our \textit{TalkingPose} model achieves state-of-the-art performance in temporal consistency and appearance preservation. This is confirmed by superior image-based metrics (SSIM, PSNR, LPIPS) reported in Table~\ref{Table:comparison_combined}, with consistent performance across the TED-talk\textsuperscript{\#}, TalkingPose, and TikTok datasets on Table~\ref{Table:comparison_TikTok}, where it also ranks second in PSNR  and LPIPS for the TED-talk\textsuperscript{\#} dataset. For video-based evaluation, \textit{TalkingPose} excels with FID-VID scores of 5.32 for TED-talk\textsuperscript{\#} and 5.07 for TalkingPose, and it obtains comparable FVD on the TikTok dataset while ranking second on the TED-talk\textsuperscript{\#} and TalkingPose datasets. Moreover, our approach effectively narrows the quality–efficiency gap relative to diffusion-based methods, handling diverse poses, gestures, and backgrounds with ease.
\cref{fig:temporal_errormap} shows two random frames from the TalkingPose set with their difference map, highlighting its stable background, consistent appearance, and accurate pose. Although \textit{AnimateAnyone} keeps appearance well via its motion module, TalkingPose yields slightly better quality and greater efficiency without specialised temporal layers or stacked‐frame training. Additional quantitative analyses are provided in the Appendix.

To assess the effectiveness of our talking avatar animation, we evaluate lip synchronization using SyncNet~\cite{chung2016out} and AKD\textsubscript{lip}, which measures alignment of lip keypoints. We conduct the evaluation on a validation set of 50 audio–video clips, each 10 seconds long, and report the results in Table~\cref{table:keypoint_distance}. As shown, our method achieves comparable SyncNet scores while attaining the best performance in AKD\textsubscript{lip}.
\subsection{Ablation Analysis}
\label{subsec:abl}
\begin{figure}[h]
  \centering
  \includegraphics[width=0.95\columnwidth]{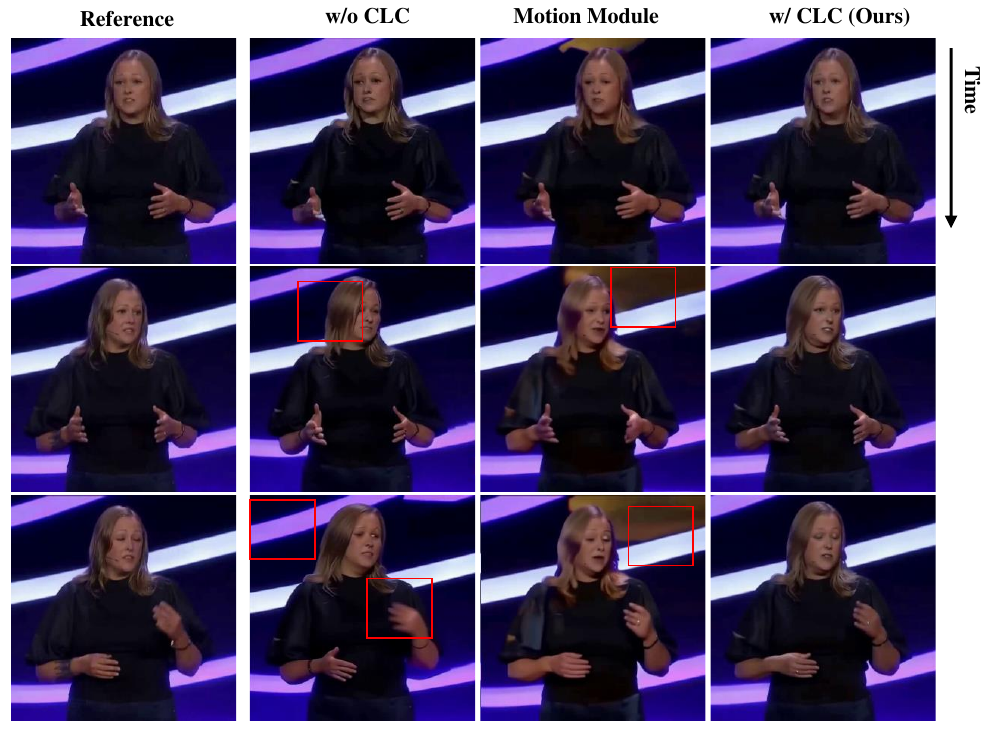}
  \captionsetup{skip=5pt} 
  \vspace{-5pt}
\caption{
\textbf{Ablation Study on Temporal Analysis}. Three sample frames from generated videos under (1) baseline without CLC, (2) with motion module, and (3) our CLC method. Red boxes mark artifacts or temporal errors vs. the reference.
\vspace{-10pt}
}
  \label{fig:ablation_pic}
\end{figure}

\begin{figure}[h]
  \centering
  \includegraphics[width=0.95\columnwidth]{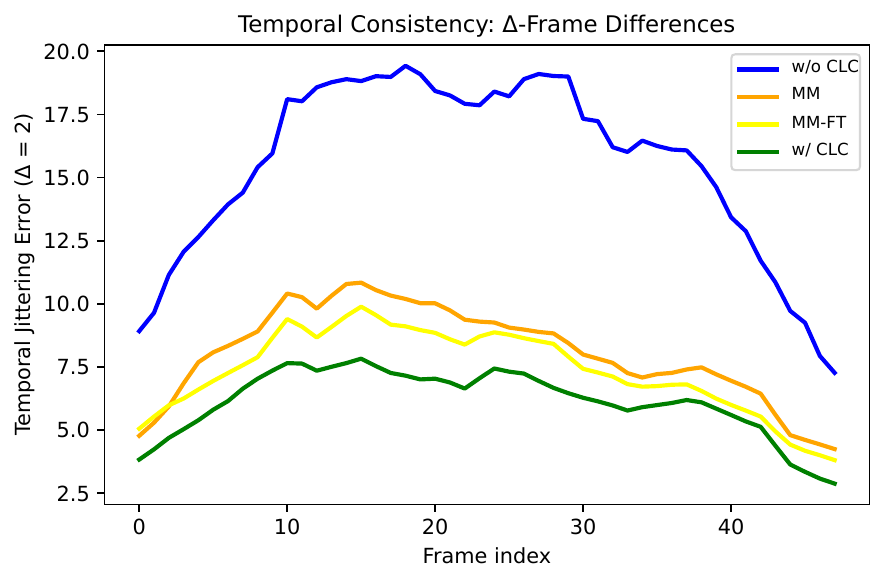}
  \captionsetup{skip=5pt} 
  \vspace{-8pt}
\caption{\textbf{Temporal Jittering Error (TJE).} ”w/o CLC” refers to the
model without a motion module, ”MM” indicates the model with a
pretrained motion module, ”MM-FT” denotes the motion module
fine-tuned, and ”Ours” represents the Base model fitted with our
CLC mechanism. Lower is better.}
  \label{fig:TJE}
\end{figure}

To assess the effectiveness of our proposed configurations, we conducted a comprehensive ablation study on the TalkingPose dataset, which includes diverse human appearances under varying lighting conditions, backgrounds, and visual attributes.
\vspace{-13pt}
\paragraph{Temporal Jittering Error (TJE).}
We introduce the \emph{Temporal Jittering Error} (TJE) metric to quantify longer-term temporal consistency. 
In addition to examining consecutive frames ($\Delta=1$), we evaluate longer-term temporal consistency by comparing pairs of frames separated by $\Delta>1$.
Concretely, let $I_t^{(\text{real})}$ and $I_{t+\Delta}^{(\text{real})}$ be two frames from the real video at time $t$ and $t+\Delta$, respectively.
Similarly, $I_t^{(\text{gen})}$ and $I_{t+\Delta}^{(\text{gen})}$ are the corresponding frames from our generated video. \\
We define:
\[
D_t^{(\text{real})} \;=\; I_{t+\Delta}^{(\text{real})} \;-\; I_t^{(\text{real})},
\quad
D_t^{(\text{gen})} \;=\; I_{t+\Delta}^{(\text{gen})} \;-\; I_t^{(\text{gen})},
\]
and measure
\[
\text{error}_t \;=\; \text{mean}\!\Bigl(\bigl| D_t^{(\text{real})} - D_t^{(\text{gen})} \bigr|\Bigr).
\]
By choosing $\Delta>1$, we capture more noticeable motion changes over a small window (e.g., 5 frames $\approx$ 200\,ms at 25\,fps), rather than relying only on single-frame intervals where slow motion might yield negligible differences.
This per-frame error is aggregated across $t=0,\ldots,N-\Delta$, providing a robust measure of how closely the generated video matches the real one over slightly longer temporal spans, thus highlighting any jitter or flicker. \\
As shown in \cref{fig:ablation_pic}, our baseline model is capable of faithfully synthesizing the character and background. However, some temporal jittering artifacts remain (highlighted with red rectangles). These artifacts are more pronounced in \cref{fig:TJE}, where our proposed CLC strategy significantly enhances temporal coherence. Quantitatively, \cref{table:comparison_ablation} shows that the base model already delivers good frame image quality, while incorporating a motion module reduces FVD from 552.82 to 285.65, it slightly degrades certain image-based metrics. This is also highlighted in \cref{fig: qualitative2} in shift contrast and artifacts in videos in supplementary. In contrast, adding our CLC method not only further reduces FVD (improving consistency beyond what the motion module achieves alone) but also yields marginal gains in image-based metrics. \\
To demonstrate CLC’s generality, we integrated it into Champ, which shares AnimateAnyone’s two-stage training. At the stage-1 checkpoint, Champ’s FVD was 580.76; with CLC ($\beta=0.05$), it fell to 276.18 on TED-Talk, confirming cross-model gains.
We then evaluated noise sampling. Switching from a fixed seed to independent $z_T$ per frame degraded FVD from 203.11 to 251.93 and LPIPS from 0.245 to 0.308, validating the fixed $z_T$ strategy in Algorithm~\ref{alg:clc}. \\
Finally, a grid search on the validation set (Table~\ref{table:beta_ablation}) showed gains above 0.1 introduce artifacts, so we select $\beta=0.05$ as the optimal stability–quality trade-off.

\begin{table}[h]
\centering
\begin{footnotesize}
\setlength{\tabcolsep}{6pt}
\begin{tabular}{lcc}
\toprule
\textbf{Method} & $\mathbf{SyncScore}_{\text{Audio}}$ & $\mathbf{AKD}_{\text{lip}}$ \\
\midrule
AnimateAnyone  & 9.11 & 6.80\\
MagicPose   & 9.89 & 7.84\\
MagicAnimate   & 10.47 & 7.95\\
MimicMotion    & 9.71 & 8.11\\
Champ   & 11.28 & 9.76\\
StableAnimator & \textbf{8.62} & 7.51\\
TalkingPose (Ours)   & 9.15 & \textbf{6.66}\\
GT             & 6.31 & --\\
\bottomrule
\end{tabular}
\end{footnotesize}
\vspace{-6pt}
\caption{Lip-sync comparison: quantitative evaluation using SyncScore and lip-only average keypoint distance (AKD\textsubscript{lip}).}
\label{table:keypoint_distance}
\end{table}

\begin{table}[h]
\centering
\begin{footnotesize}
\setlength{\tabcolsep}{3pt}
\begin{tabular}{lccccc}
\toprule
\textbf{Model} & \textbf{SSIM $\uparrow$} & \textbf{PSNR\textsubscript{float} $\uparrow$} & \textbf{LPIPS $\downarrow$} & \textbf{FID-VID $\downarrow$} & \textbf{FVD $\downarrow$} \\
\midrule

w/o CLC       & 0.747 & \textbf{21.96} & 0.190 & 26.98 & 552.82 \\
MM         & 0.696 & 20.31 & 0.219 & 8.46  & 285.65 \\
MM FT      & 0.736 & 21.87 & 0.197 & 9.56  & \textbf{192.78} \\
w/ CLC (Ours) & \textbf{0.749} & 21.94 & \textbf{0.186} & \textbf{ 5.07}  & 203.11 \\   
\bottomrule
\end{tabular}
\end{footnotesize}
\vspace{-7pt}
\caption{\textbf{Ablation on CLC.} Results on TalkingPose with different settings: “w/o CLC” is AnimateAnyone without a motion module (1st stage only), “MM” adds a pre-trained motion module, “MM-FT” fine-tunes the motion module (i.e., AnimateAnyone trained for both stages), and “w/ CLC” is AnimateAnyone with its motion module replaced by our proposed CLC.}
\label{table:comparison_ablation}
\end{table}

\begin{table}[h]
\centering
\begin{footnotesize}
\setlength{\tabcolsep}{3.8pt}
\begin{tabular}{lccccc}
\toprule
\textbf{\(\beta\) Variant} & \textbf{SSIM $\uparrow$} & \textbf{PSNR\textsubscript{float} $\uparrow$} & \textbf{LPIPS $\downarrow$} & \textbf{FID-VID $\downarrow$} & \textbf{FVD $\downarrow$} \\
\midrule
\(\beta: 0\) & 0.747 & \AJh{21.96} & \AJh{0.190} & \AJh{26.98} & \AJh{552.82} \\
\(\beta: 0.01\) & 0.748 & \textbf{21.98} & 0.188 & 8.17  & 208.50 \\
\(\beta: 0.05\) & \textbf{0.749} & 21.94 & \textbf{0.186} &\textbf{ 5.07}  & \textbf{203.11} \\
\(\beta: 0.1\)  & 0.745 & 21.84 & 0.187 & 9.52  & 206.36 \\
\(\beta: 0.2\)  & 0.625 & 16.36 & 0.269 & 23.49 & 371.84 \\
\bottomrule
\end{tabular}
\end{footnotesize}
\vspace{-7pt}
\caption{\textbf{Feedback gain (\(\beta\)) ablation study.}}
\label{table:beta_ablation}
\vspace{-15pt}
\end{table}

%% file: sec/05_conclusion.tex
\section{Conclusion}
We introduced \textit{TalkingPose}, a diffusion-based framework for animating human upper-body, including face and hand gestures, from a source appearance and driving pose sequence. Our closed-loop control mechanism ensures high temporal consistency for long video sequences. We also contribute a large-scale dataset encompassing diverse appearances, backgrounds, and gestural variations. \textit{TalkingPose} outperforms state-of-the-art methods while offering higher efficiency, paving the way for more robust and accessible human animation techniques.
\section{Acknowledgments}
\begingroup
\hyphenpenalty=50
\exhyphenpenalty=50

This work has been partially funded by the EU projects 
CORTEX\textsuperscript{2} (GA: Nr~101070192) and 
LUMINOUS (GA: Nr~101135724).

\endgroup

\vspace{-5pt}

%% file: sec/X_suppl.tex
\clearpage
\setcounter{page}{1}
\maketitlesupplementary
\begin{figure*}[t]
  \centering
  \includegraphics[width=2\columnwidth]{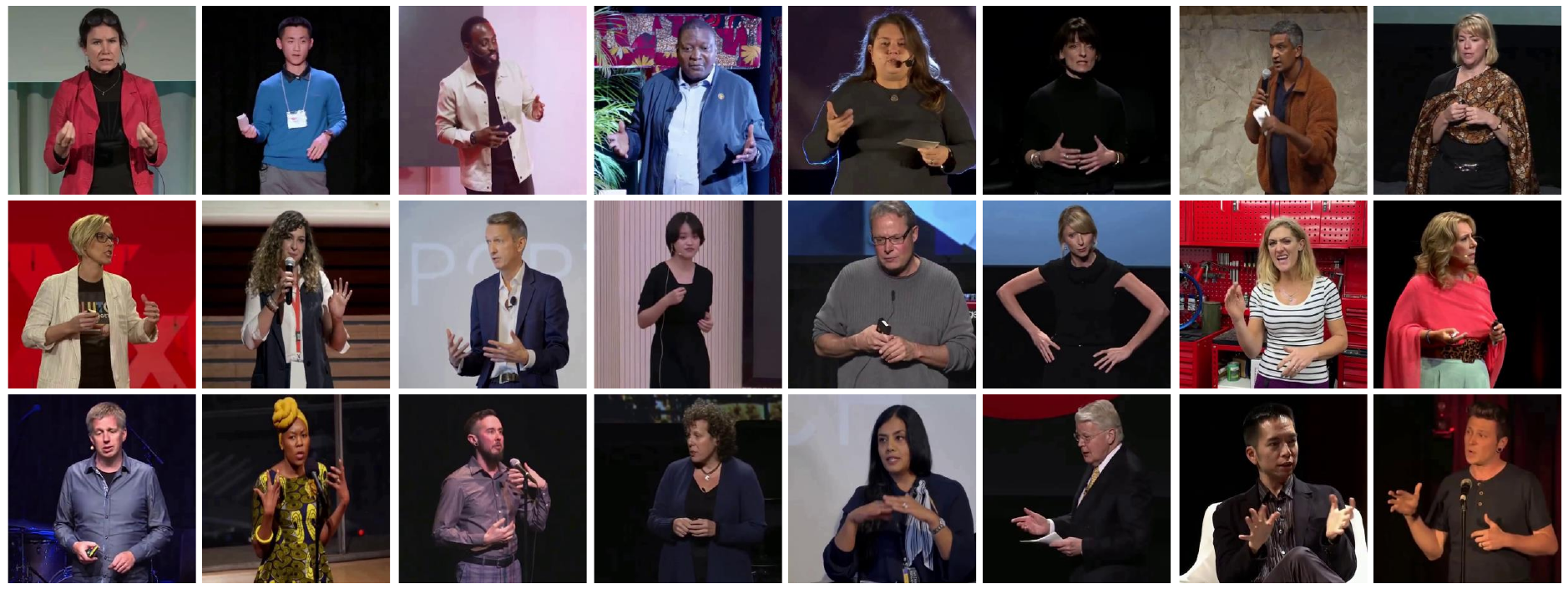}
  \captionsetup{skip=5pt} 
\vspace{-5pt}
\caption{\textbf{TalkingPose Dataset.} Example samples from the TalkingPose dataset, showcasing diversity in nationalities, age groups, and backgrounds.}
  \label{fig:supp_dataset}
\end{figure*}

This supplementary document provides additional details and insights into our work. 
In Sec.~\ref{supp:dataset}, we present a comprehensive overview of our dataset, \textit{TalkingPose}. 
%
\AJh{Sec.~\ref{supp:experiments} describes the modified TED-talk\textsuperscript{\#} dataset and further clarification regarding PSNR calculation.} Sec.~\ref{supp:analysis} provides further ablation studies and comparative analyses,
Sec.~\ref{supp:limitaions} highlights the limitations of our model and finally, Sec.~\ref{supp:addition_results} showcases additional visualizations produced by our framework.

\begin{table*}[t]
    \centering
    \resizebox{\textwidth}{!}{%
    \begin{tabular}{lcccccccccc}
    \toprule
    \textbf{Dataset} & \textbf{Real} & \textbf{Face} & \textbf{Hands} & \textbf{Full Body} & \textbf{\#Identities} & \textbf{Resolution} & \textbf{Age Range} & \textbf{Diverse BG} & \textbf{Diverse App.} \\
    \midrule
    VoxCeleb2 \cite{chung2018voxceleb2}        
       & \cmark 
       & \cmark 
       & \xmark 
       & \xmark      
       & 6k     
       & 256$\times$256                
       & Wide  
       & \cmark 
       & \cmark \\
    VFHQ \cite{xie2022vfhq}                  
       & \cmark 
       & \cmark 
       & \xmark 
       & \xmark      
       & -   
       & 512$\times$512    
       & Wide  
       & \cmark 
       & \cmark \\
    TalkingHead-1KH \cite{wang2021one} 
       & \cmark 
       & \cmark 
       & \xmark 
       & \xmark      
       & -    
       & 512$\times$512                
       & Wide  
       & \cmark 
       & \cmark \\
    AVspeech \cite{ephrat2018looking}          
       & \cmark 
       & \cmark 
       & \xmark 
       & \xmark      
       & 150k    
       & Var.                
       & Wide  
       & \cmark 
       & \cmark \\
    HDTF \cite{zhang2021flow}                  
       & \cmark 
       & \cmark 
       & \xmark 
       & \xmark      
       & 362  
       & 512$\times$512    
       & -  
       & \cmark 
       & \cmark \\
    TED-talk \cite{siarohin2021motion}           
       & \cmark 
       & \cmark 
       & \cmark 
       & \xmark 
       & 411    
       & 384$\times$384                  
       & Adult  
       & \xmark 
       & \xmark \\
    TalkShow \cite{yi2023generating}          
       & \cmark 
       & \cmark 
       & \cmark 
       & \xmark 
       & 4   
       & 1280$\times$720                  
       & Adult 
       & \xmark 
       & \xmark \\
    TikTok \cite{jafarian2021learning}              
       & \cmark 
       & \cmark 
       & \cmark 
       & partial 
       & 300  
       & Var.              
       & Young 
       & \cmark 
       & \cmark \\
       \hline
    TalkingPose (Ours)           
       & \cmark 
       & \cmark 
       & \cmark  
       & \xmark  
       & 18K  
       & 512$\times$512                
       & Wide 
       & \cmark 
       & \cmark \\
    \bottomrule
    \end{tabular}
    }
    \vspace{-6pt}
    \caption{\textbf{Comparison of face/human body datasets.} Columns indicate whether the dataset is Real (\cmark) or synthetic (\xmark), which body regions are included (partial indicates limited coverage), the approximate number of identities, total hours of video, typical resolution, gender balance, age range, and the diversity in background (BG) and appearance. Values are based on recent literature and can be adapted to each dataset’s specifics.}
    \label{tab:dataset_comparison_new}
\end{table*}

\begin{figure}[h]
  \centering
  \includegraphics[width=1\columnwidth]{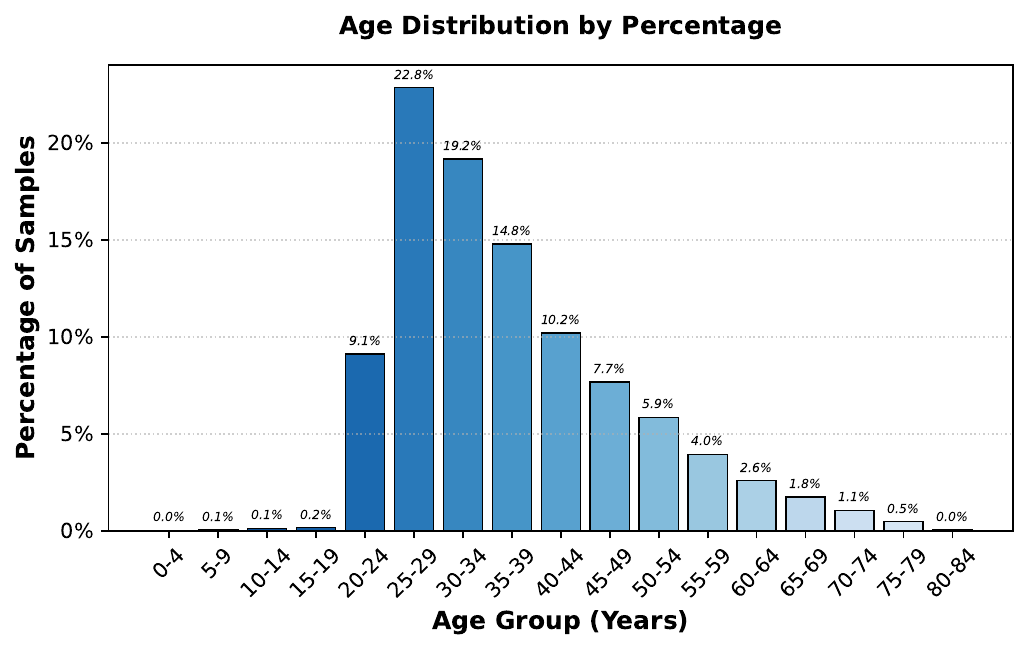}
  \captionsetup{skip=5pt} 
  \vspace{-5pt}
  \caption{\textbf{Age distribution}.}
  \label{fig:supp:age}
\end{figure}

\begin{figure}[h]
  \centering
  \includegraphics[width=1\columnwidth]{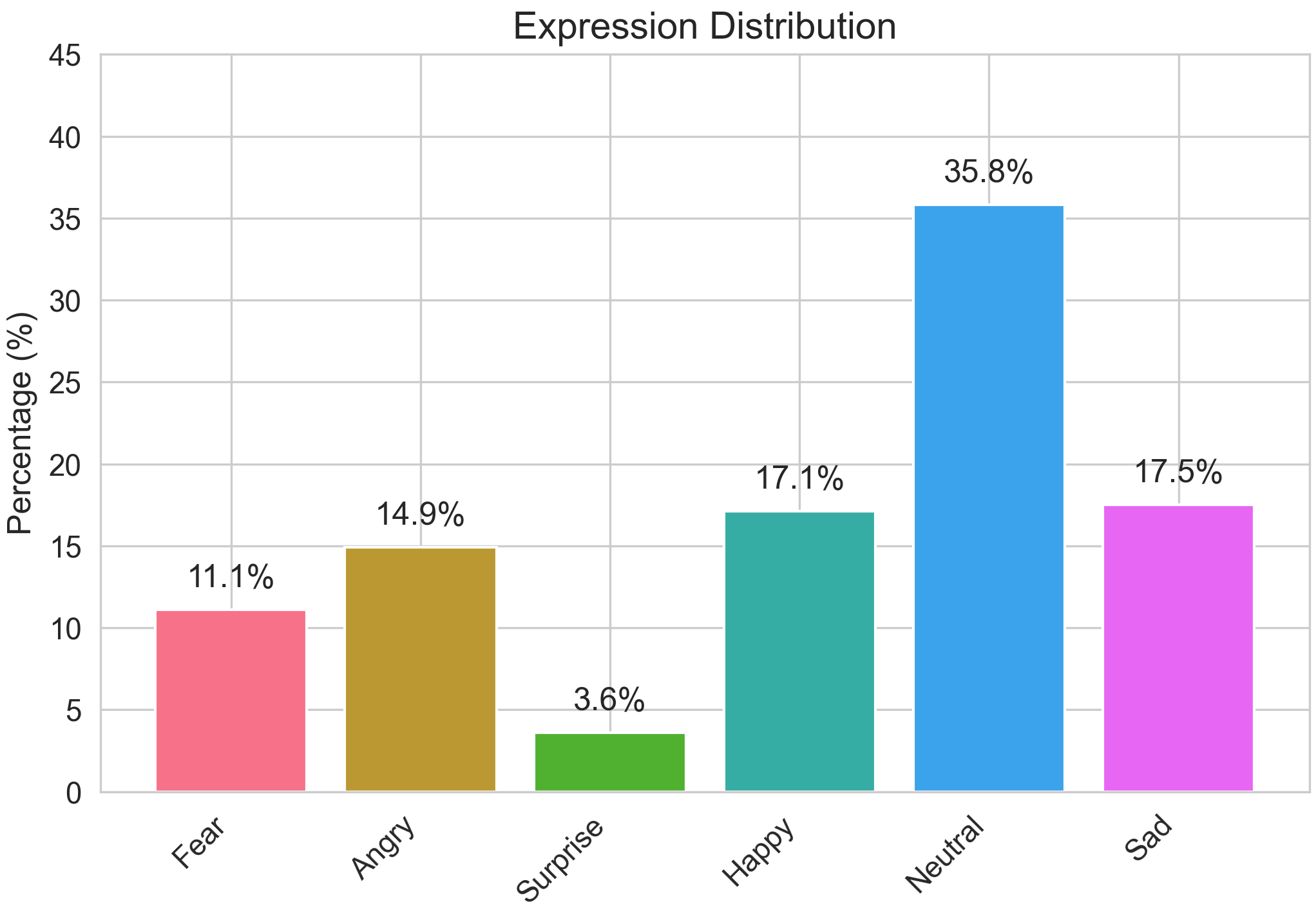}
  \captionsetup{skip=5pt} 
  \vspace{-5pt}
  \caption{\textbf{Expression distribution}.}
  \label{fig:supp:expression}
\end{figure}

\section{TalkingPose Dataset}
\label{supp:dataset}

Existing datasets~\cite{chung2018voxceleb2, wang2021one} largely fail to capture expressive upper body movements, particularly detailed facial expressions and hand gestures. The TED-Talks dataset~\cite{siarohin2021motion}, while the closest existing dataset relevant to our problem, is limited in size and lacks diversity in backgrounds, making it insufficient for studying a wide range of expressive human animations.

To address these limitations and advance research in this area, we developed the TalkingPose dataset. This dataset comprises approximately 18K video samples ---- \AJh{43× larger than TED-talk \cite{siarohin2021motion} (411 videos) and 51× larger than TikTok \cite{jafarian2021learning} (350 videos)} ---- and features individuals delivering presentations with expressive hand gestures. These videos were sourced from various YouTube channels under the "CC BY-NC-ND 4.0" license. TalkingPose represents a diverse range of individuals across different ages, genders, and backgrounds, offering a comprehensive resource for analyzing expressive upper body movements (see Fig.~\ref{fig:supp_dataset}).

\noindent \textbf{Dataset Collection and Processing Methodology:}\\
We present a detailed description of the dataset collection and processing pipeline used in our work. Our goal was to construct a large-scale, diverse dataset of human upper-body videos to support the training of robust generative models. The dataset was carefully curated to ensure a wide range of ages, genders, ethnicities, clothing styles, and environmental contexts, while maintaining strict identity separation between training and testing partitions.

\noindent \textbf{Data Acquisition and Initial Curation:}\\
We collected 21,000 raw videos from diverse YouTube sources, focusing on human presenters with visible upper-body regions. The videos were selected to include a wide variety of ages, genders, and ethnicities, as well as diverse clothing styles and background environments. This diversity is critical for training generative models to handle variations in appearance, pose, and context. Additionally, the dataset emphasizes dynamic upper-body motions, such as expressive gestures and pose variations, which are essential for synthesizing coherent and realistic avatars.

\noindent \textbf{Identity Verification and De-Duplication:}\\
To ensure identity uniqueness and prevent leakage between training and testing sets, we implemented a multi-stage identity verification pipeline. First, facial regions were extracted from each video using RetinaFace (ResNet-50 backbone) \cite{deng2019retinaface} with five-point landmark detection. Frames were filtered to retain only front-facing poses using a nose-tip alignment heuristic. Next, face embeddings were generated using ArcFace (ResNet-100) \cite{deng2019arcface}, and spectral clustering was applied to identify and remove duplicate identities across videos. A cosine similarity threshold of 0.4 was used for clustering, resulting in 18K unique identities.

\noindent \textbf{Upper-Body Processing and Quality Control:}\\
The upper-body regions were extracted using YOLO V10 \cite{wang2024yolov10} for human detection. Each frame was processed to localize the upper-body region at a resolution of 512$\times$512 pixels. Frames were discarded if the detected region after cropping is less than the desired resolution while considering the proper aspect ratio, ensuring consistent spatial proportions. The resulting dataset comprises 1250 hours of video at 20 FPS.

\noindent \textbf{Dataset Partitioning and Statistics:}\\
The curated videos were segmented into 50-frame clips for our task and the frames where the pose extraction failed were discarded. The final dataset consists of 500K clips for training (derived from 16,200 identities) and 47K clips for testing (derived from 1,800 held-out identities). This partitioning ensures that the training and testing sets are completely disjoint in terms of identities.

\noindent \textbf{Demographic Analysis:}\\
To validate the demographic diversity of our dataset, we employed the \texttt{buffalo\_l} model pack from the Insightface repository for age and gender analysis, and the DeepFace repository for expression recognition, which integrates state-of-the-art models for facial attribute analysis.
The results, illustrated in \cref{fig:supp:age} and \cref{fig:supp:expression}, confirm a balanced and varied representation of facial attributes. The expression analysis (excluding the “disgust” category) reveals that the dataset comprises six main classes: neutral (35.8\%), sad (17.5\%), happy (17.1\%), angry (14.9\%), fear (11.1\%), and surprise (3.6\%). The age distribution spans a broad range, with the majority of samples concentrated among young to middle-aged subjects. In particular, the 25--29, 30--34, and 35--39 age groups represent 22.8\%, 19.2\%, and 14.8\% of the dataset, respectively, while both the very young (0--4) and older age groups (75--84) are less represented. Furthermore, the gender distribution is balanced, with males comprising 55.4\% and females 44.6\% of the total.

This comprehensive demographic analysis demonstrates that the dataset adequately represents a wide array of facial attributes, thereby ensuring its suitability for training inclusive and generalizable facial recognition models.

\begin{table*}[!t]
\centering
\footnotesize
\setlength{\tabcolsep}{3pt}
\begin{tabular}{lcccccccccc}
\toprule
\multirow{2}{*}{\textbf{Method}} 
 & \multicolumn{4}{c}{\textbf{\AJh{TED-talk\textsuperscript{\#}}}} 
 & \multicolumn{4}{c}{\textbf{TalkingPose}} 
 & \multicolumn{2}{c}{\textbf{Model}} \\
 & {\scriptsize \textbf{AKD (face)$\downarrow$}} 
 & {\scriptsize \textbf{AKD (Hands)$\downarrow$}} 
 & {\scriptsize \textbf{AKD (torso)$\downarrow$}} 
 & {\scriptsize \textbf{CSIM}$\uparrow$} 
 & {\scriptsize \textbf{AKD (face)$\downarrow$}} 
 & {\scriptsize \textbf{AKD (Hands)$\downarrow$}} 
 & {\scriptsize \textbf{AKD (torso)$\downarrow$}} 
 & {\scriptsize \textbf{CSIM}$\uparrow$} 
 & {\scriptsize \textbf{Param. (B)$\downarrow$}} 
 & {\scriptsize \textbf{Size (GB)$\downarrow$}} \\
\midrule
AnimateAnyone~\cite{hu2024animate} 
& \underline{0.45} (0.451) & \textbf{1.45} (1.523) & \underline{3.38} (3.380) & \underline{0.54} 
& \underline{0.43} (0.436) & \textbf{1.43} (1.659) & \underline{2.67} (2.670) & 0.54 
& 2.56 & 4.77 \\

MagicPose~\cite{chang2023magicpose} 
& 0.60 (0.601) & 2.14 (2.247) & 4.48 (4.480) & 0.51 
& 0.55 (0.557) & 1.98 (2.297) & 3.49 (3.490) & 0.49 
& 2.74 & 8.53 \\

MagicAnimate~\cite{xu2024magicanimate} 
& 1.52 (1.523) & 2.99 (3.140) & 4.71 (4.710) & 0.36 
& 1.34 (1.358) & 2.53 (2.935) & 3.90 (3.900) & 0.35 
& 2.28 & 4.57 \\

MimicMotion~\cite{mimicmotion2024} 
& 2.46 (2.465) & 3.04 (3.193) & 4.78 (4.780) & 0.40 
& 1.81 (1.835) & 2.41 (2.795) & 3.32 (3.320) & 0.40 
& 2.25 & 4.20 \\

Champ~\cite{zhu2024champ} 
& 3.89 (3.898) & 5.74 (6.028) & 8.62 (8.620) & 0.27 
& 3.80 (3.853) & 5.69 (6.601) & 8.81 (8.810) & 0.29 
& 2.57 & 4.91 \\

StableAnimator~\cite{tu2024stableanimator} 
& 2.36 (2.365) & 2.93 (3.077) & 4.15 (4.150) & \textbf{0.78} 
& 1.77 (1.794) & 2.38 (2.761) & 2.85 (2.850) & \textbf{0.75} 
& 2.40 & 9.16 \\
\midrule
\textbf{TalkingPose (Ours)} 
& \textbf{0.43} (0.430) & \underline{1.46} (1.533) & \textbf{3.17} (3.170) & \underline{0.54} 
& \textbf{0.42} (0.425) & \underline{1.51} (1.751) & \textbf{2.25} (2.250) & \underline{0.56} 
& \textbf{2.10} & \textbf{3.93} \\
\bottomrule
\end{tabular}
\caption{\textbf{Quantitative comparisons on \AJh{TED-talk\textsuperscript{\#}} and TalkingPose Dataset}. The reported AKD is shown alongside its adjusted value (in parentheses), which is obtained by dividing the original AKD by the fraction of frames where the corresponding landmark is detected. Bold values indicate the best performance and underlined values indicate the second-best performance.}
\label{Table:akd_ted}
\end{table*}

\begin{table}[!t]
\centering
\footnotesize
\setlength{\tabcolsep}{2pt} 
\resizebox{0.99\linewidth}{!}{%
\begin{tabular}{@{}lcccc@{}}
\toprule
\multirow{2}{*}{\textbf{Method}} 
& \multicolumn{4}{c}{\textbf{TikTok}} \\
& {\scriptsize \textbf{AKD (face)$\downarrow$}}
& {\scriptsize \textbf{AKD (Hands)$\downarrow$}}
& {\scriptsize \textbf{AKD (torso)$\downarrow$}}
& {\scriptsize \textbf{CSIM}$\uparrow$} \\
\midrule

AnimateAnyone~\cite{hu2024animate} 
& \underline{0.63} (0.660)
& \textbf{2.91} (10.070)
& \textbf{4.60} (4.600)
& 0.45 \\

MagicPose~\cite{chang2023magicpose} 
& 0.72 (0.754)
& 4.21 (14.568)
& 8.11 (8.109)
& 0.43 \\

MagicAnimate~\cite{xu2024magicanimate} 
& 2.10 (2.200)
& 5.34 (18.478)
& 6.34 (6.340)
& 0.42 \\

MimicMotion~\cite{mimicmotion2024}  
& 3.25 (3.404)
& 5.36 (18.548)
& 6.74 (6.740)
& 0.26 \\

Champ~\cite{zhu2024champ}  
& 3.89 (4.075)
& 8.29 (28.687)
& 9.64 (9.640)
& 0.35 \\

StableAnimator~\cite{tu2024stableanimator}  
& 3.02 (3.163)
& 4.77 (16.506)
& 5.31 (5.310)
& \textbf{0.83}\textsuperscript{*} \\

\midrule
\textbf{TalkingPose (Ours)} 
& \textbf{0.61} (0.639)
& \underline{2.93} (10.139)
& \underline{4.78} (4.780)
& \underline{0.51} \\
\bottomrule
\end{tabular}%
}
\caption{\textbf{Quantitative comparisons on TikTok dataset.} Values marked with \textsuperscript{*} are reported directly from the original publications.}
\label{Table:akd_tiktok}
\vspace{-1mm}
\end{table}

\begin{table}[!t]
\centering
\footnotesize
\setlength{\tabcolsep}{4pt}      
\renewcommand{\arraystretch}{1.1} 
\resizebox{0.85\linewidth}{!}{%
\begin{tabular}{@{}lcccc@{}}
\toprule
\textbf{Method}
  & {\scriptsize $\boldsymbol{\Delta\!=\!1}$}
  & {\scriptsize $\boldsymbol{\Delta\!=\!2}$}
  & {\scriptsize $\boldsymbol{\Delta\!=\!4}$}
  & {\scriptsize $\boldsymbol{\Delta\!=\!8}$} \\
\midrule
AnimateAnyone\,w/o\,MM
  & 5.264
  & 5.882
  & 6.792
  & \underline{7.964} \\

AnimateAnyone
  & 2.997
  & \underline{4.382}
  & \underline{6.263}
  & 8.673 \\

StableAnimator
  & \underline{2.929}
  & 4.457
  & 6.440
  & 8.865 \\

\midrule
\textbf{Ours}
  & \textbf{2.336}
  & \textbf{3.366}
  & \textbf{4.677}
  & \textbf{6.197} \\
\bottomrule
\end{tabular}%
}
\caption{\textbf{Average Temporal Jittering Error (TJE)} at four frame‐offsets $\Delta$. Lower is better.}
\label{tab:tje_results}
\vspace{-4mm}
\end{table}

\begin{figure}[h]
  \centering
  \includegraphics[width=1\columnwidth]{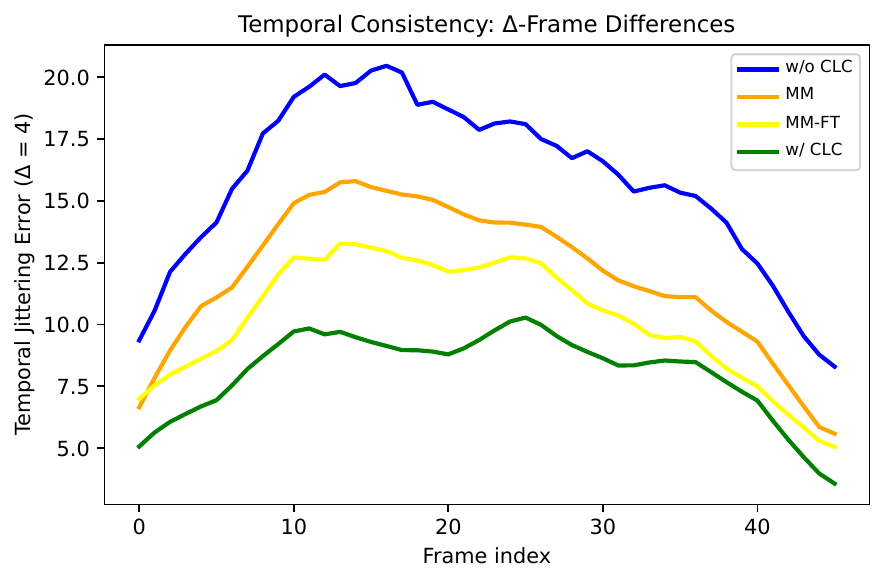}
  \captionsetup{skip=5pt} 
  \vspace{-15pt}
\caption{\textbf{Temporal Jittering Error (TJE).} ”Base” refers to the
model without a motion module, ”MM” indicates the model with a
pretrained motion module, ”MM-FT” denotes the motion module
fine-tuned, and ”Ours” represents the Base model fitted with our
CLC mechanism. Lower is better.}
\label{fig:delta4}
\end{figure}

\begin{figure}[h]
  \centering
  \includegraphics[width=1\columnwidth]{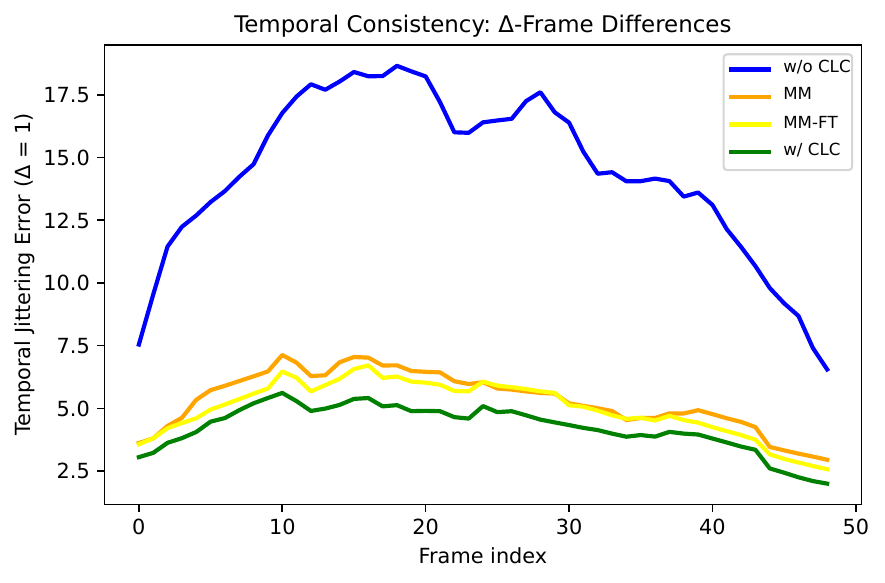}
  \captionsetup{skip=5pt} 
  \vspace{-15pt}
\caption{\textbf{Temporal Jittering Error (TJE).}}
\label{fig:delta1}
\end{figure}

\begin{figure}[h]
  \centering
  \includegraphics[width=0.95\columnwidth]{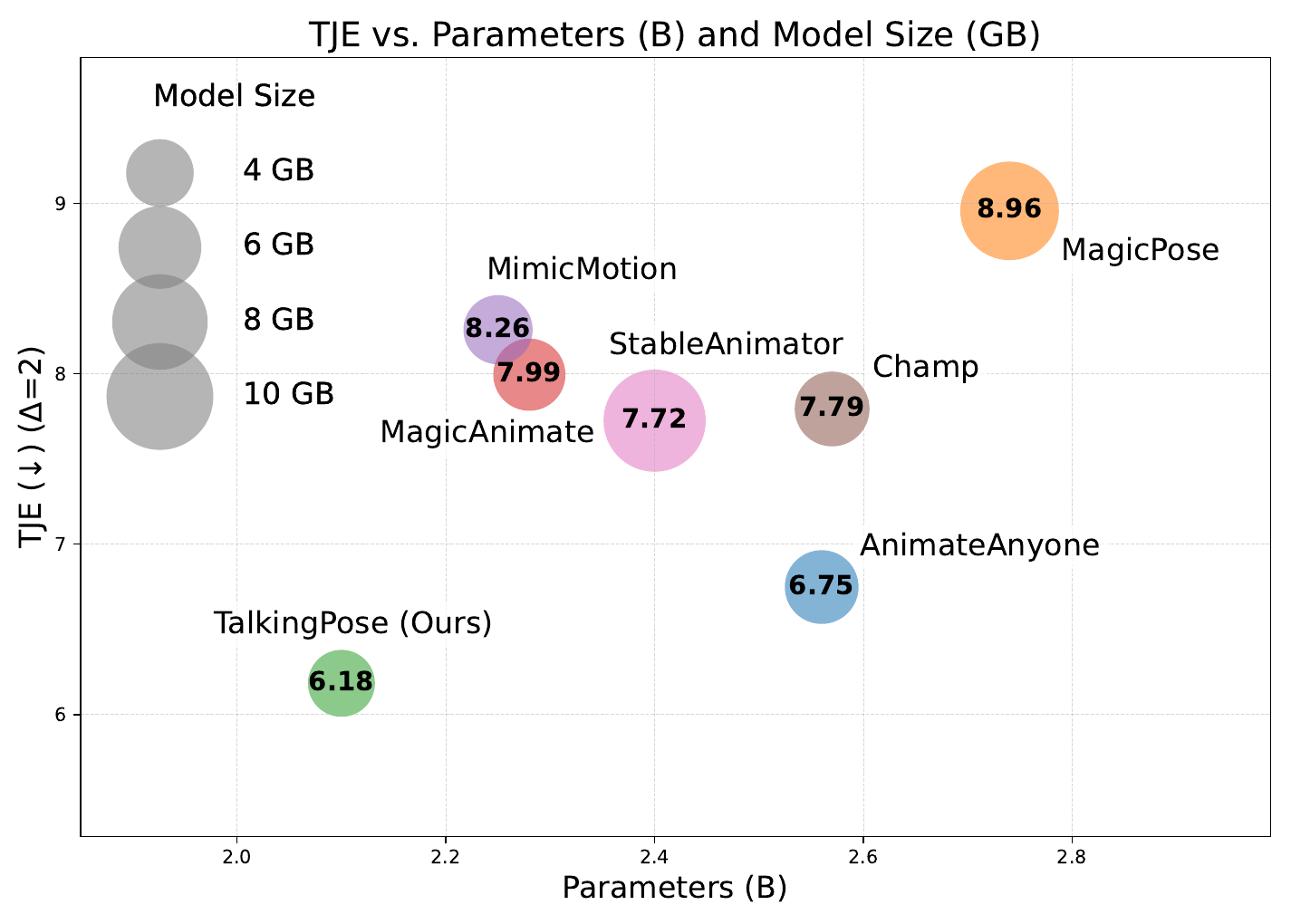}
  \captionsetup{skip=5pt} 
  \vspace{-6pt}
\caption{\AJh{\textbf{TJE vs Model Complexity.} Temporal jittering error for $\Delta=2$ compared across methods with respect to model parameters (B) and model size (GB).}}
\label{fig:TJE}
\end{figure}

\begin{figure}[h]
  \centering
  \includegraphics[width=1\columnwidth]{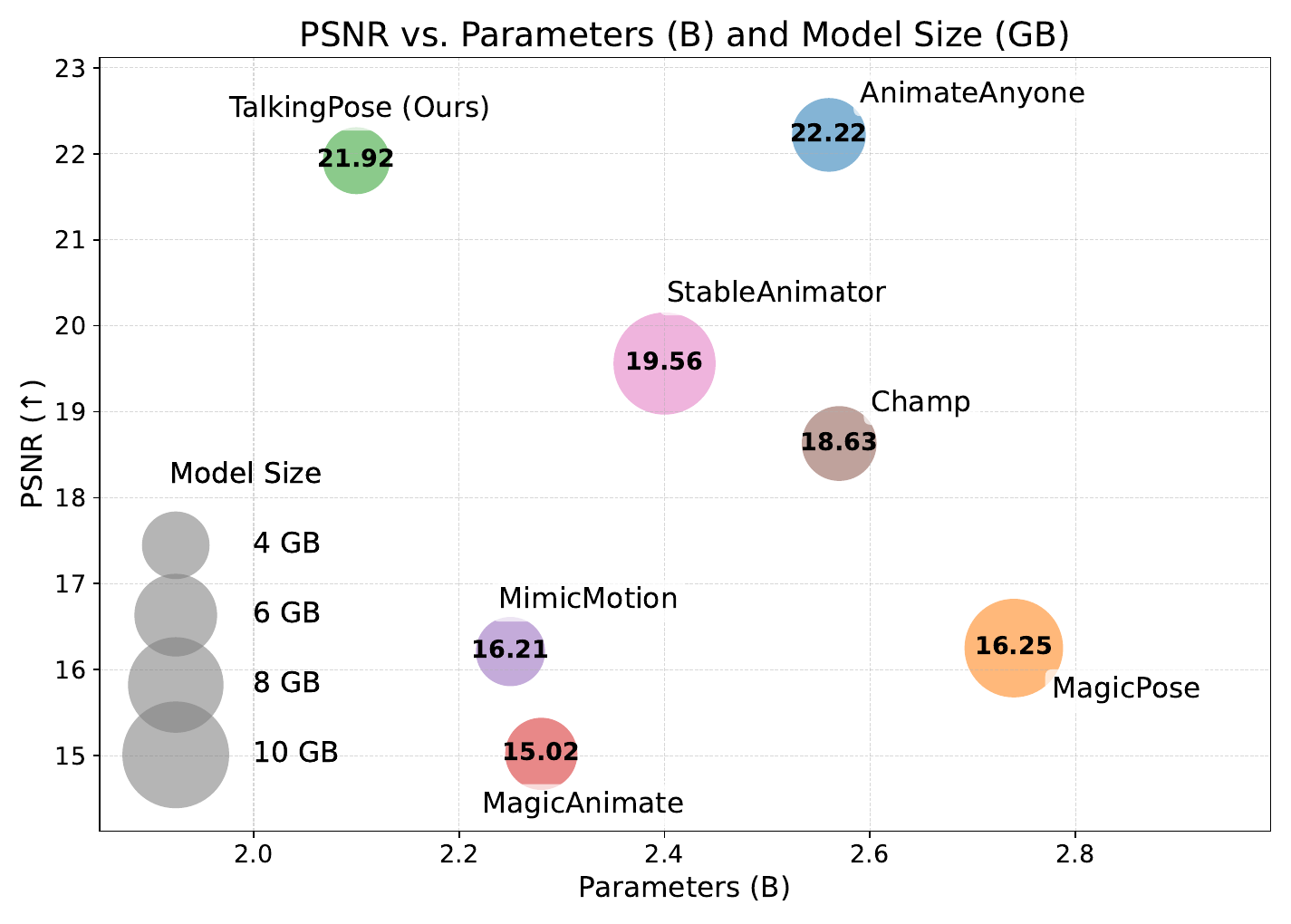}
  \captionsetup{skip=5pt} 
  \vspace{-17pt}
\caption{\AJh{\textbf{PSNR vs Model Complexity.} PSNR compared across methods with respect to model parameters (B) and model size (GB).}}
\label{fig:PSNR}
\end{figure}

\AJh{\section{Experiments}
\label{supp:experiments}
\noindent\textbf{TED-talk\textsuperscript{\#}.}
The TED-talk dataset was introduced in MRAA \cite{siarohin2021motion} (2021) using YouTube videos from the TED-Talk channel. When we attempted to download the full set of 411 videos, 62 videos were no longer downloadable. To address this, we replaced them with other videos from the same channel to maintain the total size of 411, resulting in a modified version we denote as TED-talk\textsuperscript{\#}. Importantly, all methods were evaluated on TED-talk\textsuperscript{\#} for fairness. \\
\noindent\textbf{PSNR calculation.} The discrepancy in reported PSNR values across prior works arises from differences in data type handling when computing Mean Squared Error (MSE) for PSNR calculation. Specifically, using integer values can cause numerical overflow. This issue originates from the Disco \cite{wang2024disco} evaluation toolkit, which was widely used in earlier papers \cite{hu2024animate,chang2023magicpose,zhu2024champ}, and is also observable when comparing different versions of the Disco paper \cite{wang2024disco} (e.g., v1 on arXiv vs. v3). In the updated Disco toolkit, MSE is computed using floating point, and recent works (e.g., \cite{mimicmotion2024,xu2024magicanimate,tu2024stableanimator}) follow this update. StableAnimator \cite{tu2024stableanimator} also reported both cases for comparison.}

\section{Additional Analysis}
\label{supp:analysis}
In addition to the image-based metrics (SSIM, PSNR, LPIPS) and temporal evaluations (FID-VID, FVD), we further measure performance using the Average Keypoint Distance (AKD) \cite{JMLR:v12:gashler11a} computed with MediaPipe \cite{lugaresi2019mediapipe} and the face-based Cosine Similarity (CSIM) \cite{richardson2021encoding,huang2020curricularface}, following \cite{chang2023magicpose}. During experimentation, we observed that AKD can be inaccurate for frames with noticeable artifacts, because the landmark extractor often fails to detect landmarks in those frames.
To address this, we first calculate AKD only on frames deemed valid across all methods (i.e., where faces, hands, or torsos are successfully detected). We then compute the average percentage of valid landmarks across all frames—so frames with artifacts that cause the extractor to fail for one method but not for others result in a lower detection fraction for that particular method. Finally, we divide the original AKD values by this detection fraction to achieve a fairer comparison.
\begin{figure*}[t]
  \centering
  \includegraphics[width=2\columnwidth]{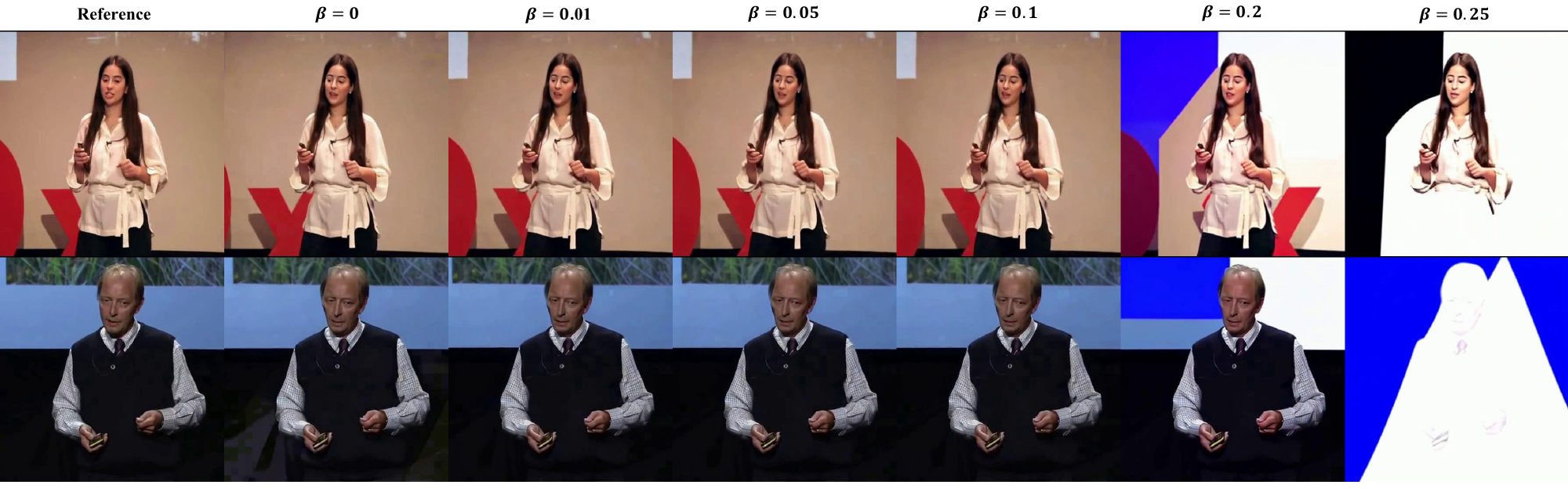}
  \captionsetup{skip=5pt} 
\vspace{-5pt}
\caption{\AJh{\textbf{Ablation study on $\beta$ values.}}}
  \label{fig:beta}
\end{figure*}
Tables~\ref{Table:akd_ted} and \ref{Table:akd_tiktok} show that \textit{TalkingPose} outperforms all baselines on most metrics, ranking second in AKD for hands and CSIM, indicating strong identity preservation. Moreover, our method is the most efficient: it adds no additional temporal layers and instead relies on a more effective CLC mechanism, resulting in fewer parameters and reduced computational overhead. Table~\ref{Table:akd_ted} further quantifies this efficiency: because the model does not include temporal parameters in its inference-time size, does not require a second training stage as in \cite{hu2024animate}, and does not train on video stacks of frames like StableAnimator \cite{tu2024stableanimator} or MimicMotion \cite{mimicmotion2024}, its GPU memory requirement at inference is minimal—only one frame is held in memory at a time—whereas all other methods operate on stacks of frames. \\
To further analyze temporal consistency, we performed a temporal jittering error (TJE) evaluation across several $\Delta$ values (e.g., $\Delta{=}1$, \cref{fig:delta1}; $\Delta{=}4$, \cref{fig:delta4}). These plots show that our CLC mechanism consistently outperforms the motion module and, by a large margin, the base model, indicating strong temporal stability. \\
We additionally evaluated long videos (approximately 1k frames per sample) on a 50-video validation set from TED-talk\textsuperscript{\#}. We compared our method with AnimateAnyone \cite{hu2024animate} (a Stable Diffusion–based \cite{rombach2022high} method with a motion module) and StableAnimator \cite{tu2024stableanimator}(based on Stable Video Diffusion \cite{tu2024stableanimator}). Although these methods maintain reasonable temporal consistency over short durations, performance degrades on long sequences because concatenating short generated chunks introduces temporal drift, as reported in \cref{tab:tje_results}. Our method outperforms these approaches while also being significantly more efficient.
\AJh{Inspired by the representation style in \cite{liu2025edcflow}, we plot the Temporal Jittering Error (TJE) against model parameters and size in \cref{fig:TJE}. TJE measures the temporal consistency of generated videos with respect to the ground truth in same-identity animation on the TED-talk\textsuperscript{\#} dataset. In \cref{fig:PSNR}, we report the PSNR, which evaluates the photorealism of the generated frames. As shown, our proposed method consistently outperforms competing approaches in terms of TJE, and achieves the highest PSNR among all methods, performing on par with AnimateAnyone, the strongest baseline. Importantly, these improvements are obtained in a highly efficient manner: our CLC-based approach requires neither additional parameters to enforce temporal consistency nor training on temporal data, resulting in a significantly smaller and more parameter-efficient model. \\
To further demonstrate the effectiveness of our proposed CLC, beyond the quantitative ablation on $\beta$ values, we present two examples from the TalkingPose dataset in \cref{fig:beta}. When $\beta=0$, the generated results exhibit high visual quality, but temporal consistency is not maintained. For instance, in the first row, the background is slightly misaligned compared to the ground truth. Such artifacts become more pronounced in videos, manifesting as jittering. Notably, even a small increase in $\beta$, which activates the feedback loop, can effectively suppress these artifacts. On the other hand, setting $\beta$ too high (e.g., $\beta>0.1$) introduces significant distortions into the output. Through our experiments, we found the optimal value to be $\beta=0.05$.}

\section{Limitations}
\label{supp:limitaions}
\begin{figure}[h]
  \centering
  \includegraphics[width=1\columnwidth]{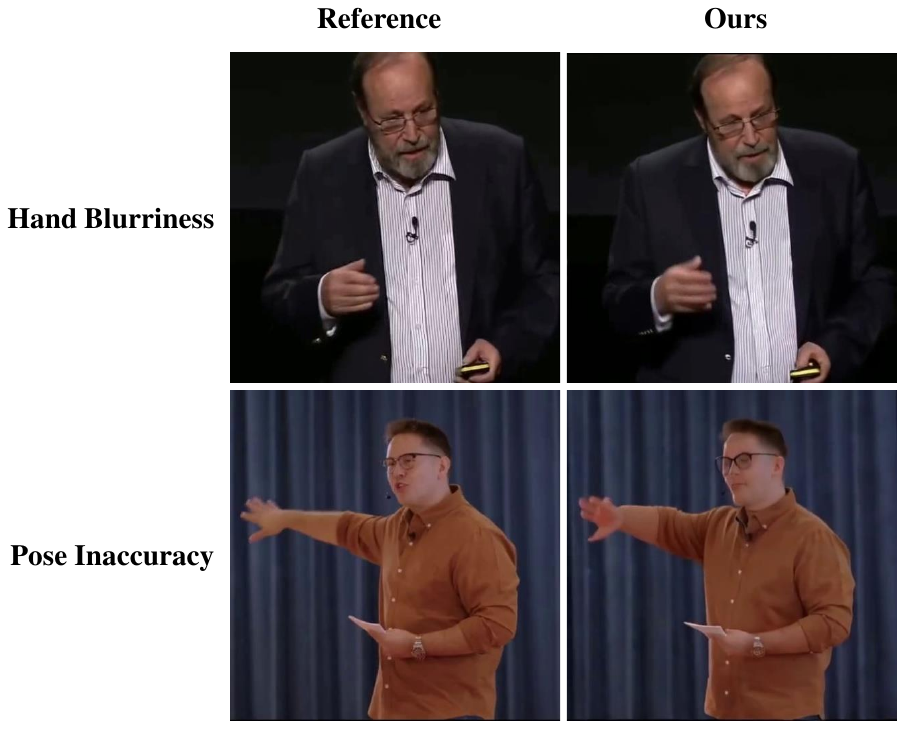}
  \captionsetup{skip=5pt} 
  \vspace{-15pt}
\caption{\textbf{Limitations.} 
}
\label{fig:limit}
\end{figure}

Our framework exhibits certain constraints when confronted with extreme poses, particularly those involving head or torso rotations beyond 90 degrees. In such scenarios, the accurate extraction of 3D poses becomes challenging (as illustrated in the top section of Fig.~\ref{fig:limit}). Additionally, the system's performance is compromised when the initial source image fails to adequately represent or partially occludes the hand. Consequently, in some instances, the character's hand may appear slightly blurred, stemming from the diffusion backbone's inability to faithfully generate the hand region. These limitations highlight areas for potential future improvements in our approach.
Furthermore, the strong generative prior of stable diffusion can sometimes cause the framework to hallucinate certain aspects of the appearance in the generated motion, leading to inaccuracies.
\section{Additional Results}
\label{supp:addition_results}
\begin{figure*}[t]
  \centering
  \includegraphics[width=2.03\columnwidth]{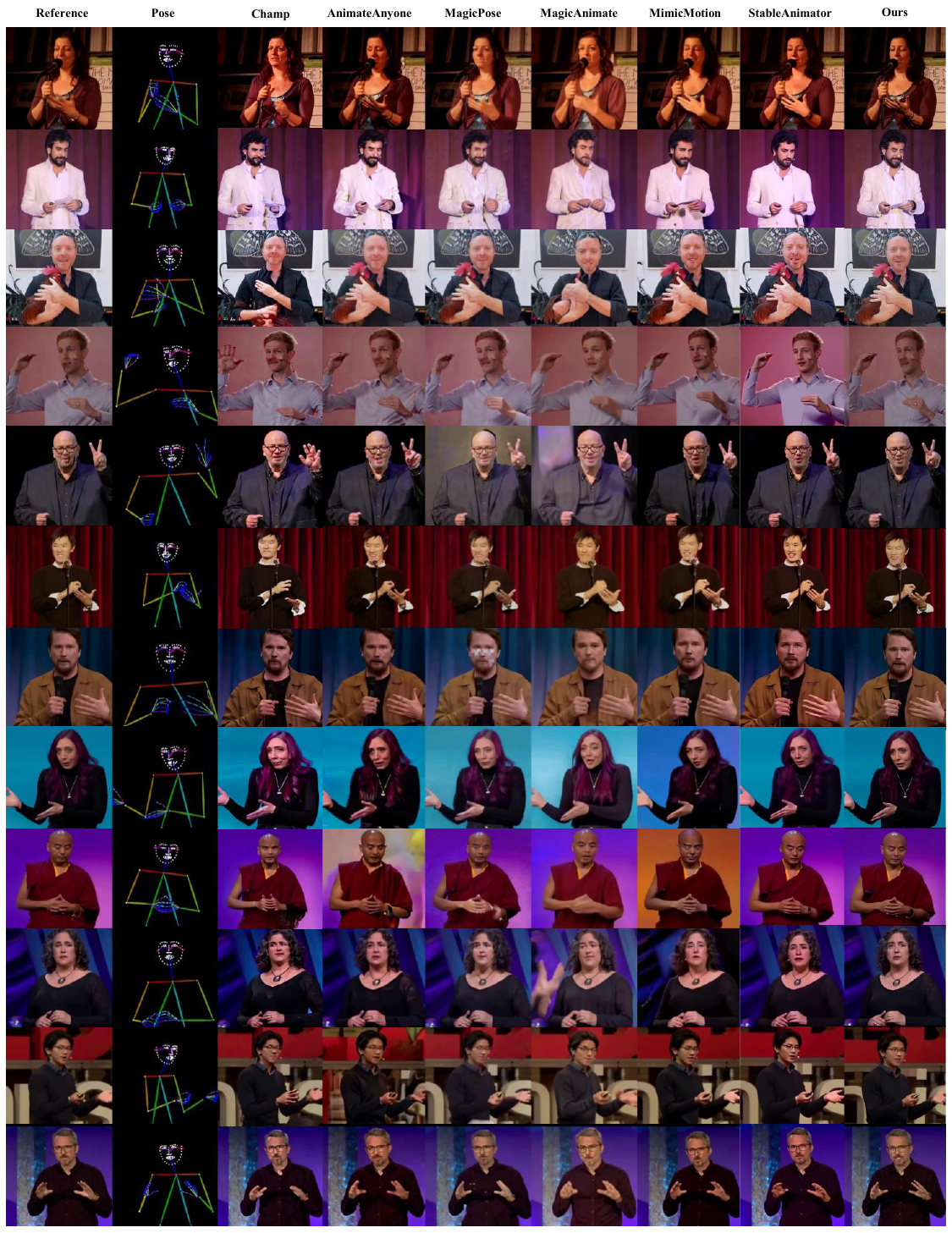}
  \captionsetup{skip=5pt} 
\vspace{-4pt}
\caption{\textbf{Qualitative Comparison}. Comparison of our framework with the state-of-the-art methods Champ \cite{zhu2024champ}, AnimateAnyone \cite{hu2024animate}, MagicPose \cite{chang2023magicpose}, MagicAnimate \cite{xu2024magicanimate}, MimicMotion \cite{mimicmotion2024} and StableAnimator \cite{tu2024stableanimator} on the TalkingPose dataset (top six rows) and the TED-talk\textsuperscript{\#} dataset (bottom six rows).}
  \label{fig:comparision_ted}
\end{figure*}

\begin{figure*}[t]
  \centering
  \includegraphics[width=2.03\columnwidth]{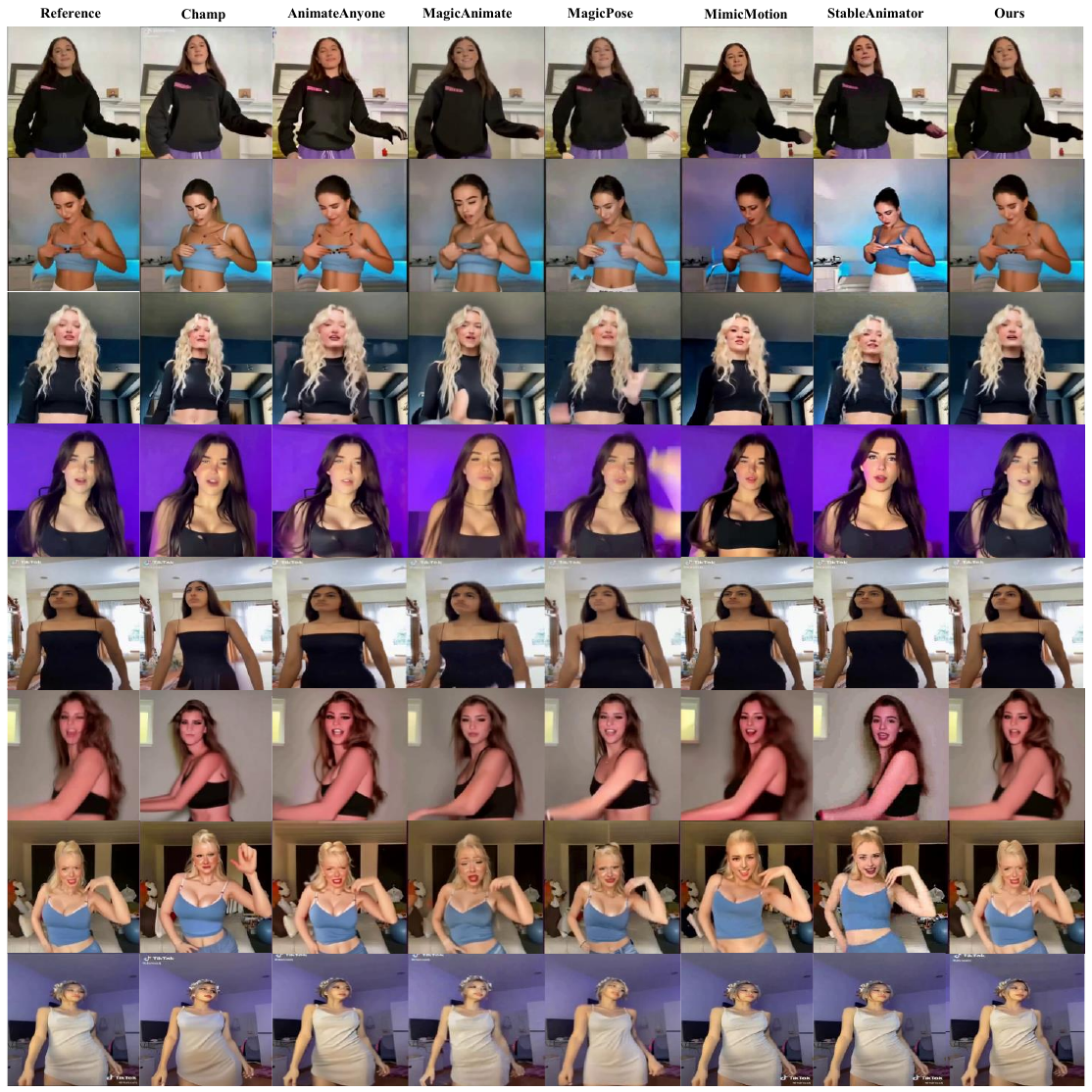}
  \captionsetup{skip=5.5pt} 
\vspace{-5pt}
\caption{\textbf{Qualitative Comparison}. Comparison of our framework with the state-of-the-art methods Champ \cite{zhu2024champ}, AnimateAnyone \cite{hu2024animate}, MagicPose \cite{chang2023magicpose}, MagicAnimate \cite{xu2024magicanimate}, MimicMotion \cite{mimicmotion2024} and StableAnimator \cite{tu2024stableanimator} on the TikTok dataset.}
  \label{fig:comparision_tiktok}
\end{figure*}

\begin{figure*}[t]
  \centering
  \includegraphics[width=2\columnwidth]{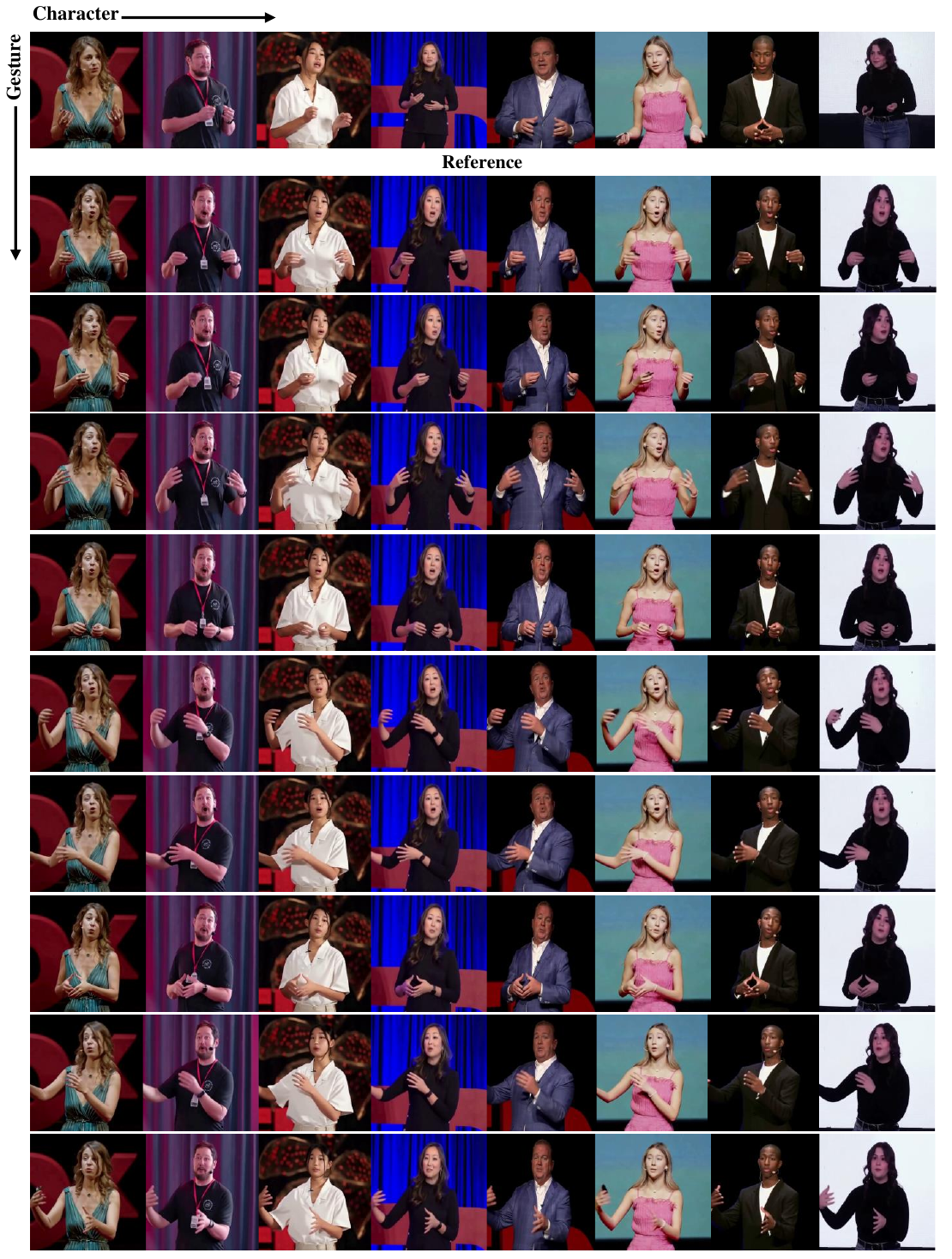}
  \captionsetup{skip=5pt} 
\vspace{-5pt}
\caption{
\textbf{Cross-identity Animation}. The rows represent the \textit{same pose} with different identities while the columns represent the different poses within the \textit{same identity}. 
}
  \label{fig:cross_animation}
\end{figure*}

We present additional results in Fig.~\ref{fig:comparision_ted} and Fig.~\ref{fig:comparision_tiktok} to further demonstrate the effectiveness of our framework. 
Additionally, Fig.~\ref{fig:cross_animation} highlights our framework's ability to replicate the same pose across multiple identities with high precision, regardless of variations in age, ethnicity, or gender.
We encourage readers to refer to the supplementary video for additional visualizations.

%% file: main.bib
@String(CVPR= {IEEE Conf. Comput. Vis. Pattern Recog.})

@String(ICLR = {Int. Conf. Learn. Represent.})

@String(IJCAI = {IJCAI})

@String(AAAI = {AAAI})

@String(CVPRW= {IEEE Conf. Comput. Vis. Pattern Recog. Worksh.})

@String(CVPR  = {CVPR})

@String(ICLR  = {ICLR})

@String(CVPRW= {CVPRW})

@inproceedings{wang2021one,
  title={One-shot free-view neural talking-head synthesis for video conferencing},
  author={Wang, Ting-Chun and Mallya, Arun and Liu, Ming-Yu},
  booktitle={Proceedings of the IEEE/CVF Conference on Computer Vision and Pattern Recognition},
  pages={10039--10049},
  year={2021}
}

@article{siarohin2019first,
  title={First order motion model for image animation},
  author={Siarohin, Aliaksandr and Lathuili{\`e}re, St{\'e}phane and Tulyakov, Sergey and Ricci, Elisa and Sebe, Nicu},
  journal={Advances in Neural Information Processing Systems},
  volume={32},
  year={2019}
}

@InProceedings{Thies_2016_CVPR,
author = {Thies, Justus and Zollhofer, Michael and Stamminger, Marc and Theobalt, Christian and Niessner, Matthias},
title = {Face2Face: Real-Time Face Capture and Reenactment of RGB Videos},
booktitle = {Proceedings of the IEEE Conference on Computer Vision and Pattern Recognition},
month = {June},
year = {2016}
}

@inproceedings{ronneberger2015u,
  title={U-net: Convolutional networks for biomedical image segmentation},
  author={Ronneberger, Olaf and Fischer, Philipp and Brox, Thomas},
  booktitle={Medical Image Computing and Computer-Assisted Intervention--MICCAI 2015: 18th International Conference, Munich, Germany, October 5-9, 2015, Proceedings, Part III 18},
  pages={234--241},
  year={2015},
  organization={Springer}
}

@article{mildenhall2021nerf,
  title={Nerf: Representing scenes as neural radiance fields for view synthesis},
  author={Mildenhall, Ben and Srinivasan, Pratul P and Tancik, Matthew and Barron, Jonathan T and Ramamoorthi, Ravi and Ng, Ren},
  journal={Communications of the ACM},
  volume={65},
  number={1},
  pages={99--106},
  year={2021},
  publisher={ACM New York, NY, USA}
}

@article{JMLR:v12:gashler11a,
  author  = {Michael Gashler},
  title   = {Waffles: A Machine Learning Toolkit},
  journal = {Journal of Machine Learning Research},
  year    = {2011},
  volume  = {12},
  number  = {69},
  pages   = {2383--2387}}

@inproceedings{
wang2022latent,
title={Latent Image Animator: Learning to Animate Images via Latent Space Navigation},
author={Yaohui Wang and Di Yang and Francois Bremond and Antitza Dantcheva},
booktitle={International Conference on Learning Representations},
year={2022}
}

@article{ho2020denoising,
  title={Denoising diffusion probabilistic models},
  author={Ho, Jonathan and Jain, Ajay and Abbeel, Pieter},
  journal={Advances in Neural Information Processing Systems},
  volume={33},
  pages={6840--6851},
  year={2020}
}

@article{xu2024vasa,
  title={VASA-1: Lifelike Audio-Driven Talking Faces Generated in Real Time},
  author={Xu, Sicheng and Chen, Guojun and Guo, Yu-Xiao and Yang, Jiaolong and Li, Chong and Zang, Zhenyu and Zhang, Yizhong and Tong, Xin and Guo, Baining},
  journal={arXiv preprint arXiv:2404.10667},
  year={2024}
}

@article{tian2024emo,
  title={EMO: Emote Portrait Alive-Generating Expressive Portrait Videos with Audio2Video Diffusion Model under Weak Conditions},
  author={Tian, Linrui and Wang, Qi and Zhang, Bang and Bo, Liefeng},
  journal={arXiv preprint arXiv:2402.17485},
  year={2024}
}

@article{thies2015real,
  title={Real-time expression transfer for facial reenactment.},
  author={Thies, Justus and Zollh{\"o}fer, Michael and Nie{\ss}ner, Matthias and Valgaerts, Levi and Stamminger, Marc and Theobalt, Christian},
  journal={ACM Trans. Graph.},
  volume={34},
  number={6},
  pages={183--1},
  year={2015}
}

@inproceedings{prinzler2023diner,
  title={Diner: Depth-aware image-based neural radiance fields},
  author={Prinzler, Malte and Hilliges, Otmar and Thies, Justus},
  booktitle={Proceedings of the IEEE/CVF Conference on Computer Vision and Pattern Recognition},
  pages={12449--12459},
  year={2023}
}

@inproceedings{zielonka2023instant,
  title={Instant volumetric head avatars},
  author={Zielonka, Wojciech and Bolkart, Timo and Thies, Justus},
  booktitle={Proceedings of the IEEE/CVF Conference on Computer Vision and Pattern Recognition},
  pages={4574--4584},
  year={2023}
}

@inproceedings{liu2019soft,
  title={Soft rasterizer: A differentiable renderer for image-based 3d reasoning},
  author={Liu, Shichen and Li, Tianye and Chen, Weikai and Li, Hao},
  booktitle={Proceedings of the IEEE/CVF International Conference on Computer Vision},
  pages={7708--7717},
  year={2019}
}

@article{kerbl20233d,
  title={3{D} Gaussian Splatting for Real-Time Radiance Field Rendering.},
  author={Kerbl, Bernhard and Kopanas, Georgios and Leimk{\"u}hler, Thomas and Drettakis, George},
  journal={ACM Trans. Graph.},
  volume={42},
  number={4},
  pages={139--1},
  year={2023}
}

@inproceedings{xu2024gaussian,
  title={Gaussian head avatar: Ultra high-fidelity head avatar via dynamic gaussians},
  author={Xu, Yuelang and Chen, Benwang and Li, Zhe and Zhang, Hongwen and Wang, Lizhen and Zheng, Zerong and Liu, Yebin},
  booktitle={Proceedings of the IEEE/CVF Conference on Computer Vision and Pattern Recognition},
  pages={1931--1941},
  year={2024}
}

@article{goodfellow2014generative,
  title={Generative adversarial nets},
  author={Goodfellow, Ian and Pouget-Abadie, Jean and Mirza, Mehdi and Xu, Bing and Warde-Farley, David and Ozair, Sherjil and Courville, Aaron and Bengio, Yoshua},
  journal={Advances in Neural Information Processing Systems},
  volume={27},
  year={2014}
}

@article{dhariwal2021diffusion,
  title={Diffusion models beat gans on image synthesis},
  author={Dhariwal, Prafulla and Nichol, Alexander},
  journal={Advances in Neural Information Processing Systems},
  volume={34},
  pages={8780--8794},
  year={2021}
}

@inproceedings{rombach2022high,
  title={High-resolution image synthesis with latent diffusion models},
  author={Rombach, Robin and Blattmann, Andreas and Lorenz, Dominik and Esser, Patrick and Ommer, Bj{\"o}rn},
  booktitle={Proceedings of the IEEE/CVF conference on computer vision and pattern recognition},
  pages={10684--10695},
  year={2022}
}

@article{ho2022video,
  title={Video diffusion models},
  author={Ho, Jonathan and Salimans, Tim and Gritsenko, Alexey and Chan, William and Norouzi, Mohammad and Fleet, David J},
  journal={Advances in Neural Information Processing Systems},
  volume={35},
  pages={8633--8646},
  year={2022}
}

@article{ma2024follow,
  title={Follow-Your-Emoji: Fine-Controllable and Expressive Freestyle Portrait Animation},
  author={Ma, Yue and Liu, Hongyu and Wang, Hongfa and Pan, Heng and He, Yingqing and Yuan, Junkun and Zeng, Ailing and Cai, Chengfei and Shum, Heung-Yeung and Liu, Wei and others},
  journal={arXiv preprint arXiv:2406.01900},
  year={2024}
}

@inproceedings{hu2024animate,
  title={Animate anyone: Consistent and controllable image-to-video synthesis for character animation},
  author={Hu, Li},
  booktitle={Proceedings of the IEEE/CVF Conference on Computer Vision and Pattern Recognition},
  pages={8153--8163},
  year={2024}
}

@inproceedings{chang2023magicpose,
  title={MagicPose: Realistic Human Poses and Facial Expressions Retargeting with Identity-aware Diffusion},
  author={Chang, Di and Shi, Yichun and Gao, Quankai and Xu, Hongyi and Fu, Jessica and Song, Guoxian and Yan, Qing and Zhu, Yizhe and Yang, Xiao and Soleymani, Mohammad},
  booktitle={Forty-first International Conference on Machine Learning},
  year={2023}
}

@inproceedings{xu2024magicanimate,
  title={Magicanimate: Temporally consistent human image animation using diffusion model},
  author={Xu, Zhongcong and Zhang, Jianfeng and Liew, Jun Hao and Yan, Hanshu and Liu, Jia-Wei and Zhang, Chenxu and Feng, Jiashi and Shou, Mike Zheng},
  booktitle={Proceedings of the IEEE/CVF Conference on Computer Vision and Pattern Recognition},
  pages={1481--1490},
  year={2024}
}

@book{kailath1980linear,
  title={Linear Systems},
  author={Kailath, T.},
  isbn={9780135369616},
  lccn={79014928},
  series={Information and System Sciences Series},
  url={https://books.google.de/books?id=ggYqAQAAMAAJ},
  year={1980},
  publisher={Prentice-Hall}
}

@inproceedings{siarohin2021motion,
  title={Motion representations for articulated animation},
  author={Siarohin, Aliaksandr and Woodford, Oliver J and Ren, Jian and Chai, Menglei and Tulyakov, Sergey},
  booktitle={Proceedings of the IEEE/CVF Conference on Computer Vision and Pattern Recognition},
  pages={13653--13662},
  year={2021}
}

@article{FLAME:SiggraphAsia2017, 
  title = {Learning a model of facial shape and expression from {4D} scans}, 
  author = {Li, Tianye and Bolkart, Timo and Black, Michael. J. and Li, Hao and Romero, Javier}, 
  journal = {ACM Transactions on Graphics, (Proc. SIGGRAPH Asia)}, 
  volume = {36}, 
  number = {6}, 
  year = {2017}, 
  pages = {194:1--194:17},
  url = {https://doi.org/10.1145/3130800.3130813} 
}

@article{SMPL:2015,
      author = {Loper, Matthew and Mahmood, Naureen and Romero, Javier and Pons-Moll, Gerard and Black, Michael J.},
      title = {{SMPL}: A Skinned Multi-Person Linear Model},
      journal = {ACM Trans. Graphics (Proc. SIGGRAPH Asia)},
      month = oct,
      number = {6},
      pages = {248:1--248:16},
      publisher = {ACM},
      volume = {34},
      year = {2015}
    }

@article{sitzmann2020implicit,
  title={Implicit neural representations with periodic activation functions},
  author={Sitzmann, Vincent and Martel, Julien and Bergman, Alexander and Lindell, David and Wetzstein, Gordon},
  journal={Advances in Neural Information Processing Systems},
  volume={33},
  pages={7462--7473},
  year={2020}
}

@inproceedings{yang2022face2face,
  title={Face2face $\rho$: Real-time high-resolution one-shot face reenactment},
  author={Yang, Kewei and Chen, Kang and Guo, Daoliang and Zhang, Song-Hai and Guo, Yuan-Chen and Zhang, Weidong},
  booktitle={European Conference on Computer Vision},
  pages={55--71},
  year={2022},
  organization={Springer}
}

@article{lin2024cyberhost,
  title={CyberHost: Taming Audio-driven Avatar Diffusion Model with Region Codebook Attention},
  author={Lin, Gaojie and Jiang, Jianwen and Liang, Chao and Zhong, Tianyun and Yang, Jiaqi and Zheng, Yanbo},
  journal={arXiv preprint arXiv:2409.01876},
  year={2024}
}

@article{jiang2024loopy,
title={Loopy: Taming Audio-Driven Portrait Avatar with Long-Term Motion Dependency},
author={Jiang, Jianwen and Liang, Chao and Yang, Jiaqi and Lin, Gaojie and Zhong, Tianyun and Zheng, Yanbo},
journal={arXiv preprint arXiv:2409.02634},
year={2024}
}

@inproceedings{mimicmotion2024,
  title={MimicMotion: High-Quality Human Motion Video Generation with Confidence-aware Pose Guidance},
  author={Yuang Zhang and Jiaxi Gu and Li-Wen Wang and Han Wang and Junqi Cheng and Yuefeng Zhu and Fangyuan Zou},
  booktitle={International Conference on Machine Learning},
  year={2025}
}

@article{kim2024tcan,
  title={TCAN: Animating Human Images with Temporally Consistent Pose Guidance using Diffusion Models},
  author={Kim, Jeongho and Kim, Min-Jung and Lee, Junsoo and Choo, Jaegul},
  journal={arXiv preprint arXiv:2407.09012},
  year={2024}
}

@article{zhu2024champ,
  title={Champ: Controllable and consistent human image animation with 3d parametric guidance},
  author={Zhu, Shenhao and Chen, Junming Leo and Dai, Zuozhuo and Xu, Yinghui and Cao, Xun and Yao, Yao and Zhu, Hao and Zhu, Siyu},
  journal={arXiv preprint arXiv:2403.14781},
  year={2024}
}

@article{ho2022cascaded,
  title={Cascaded diffusion models for high fidelity image generation},
  author={Ho, Jonathan and Saharia, Chitwan and Chan, William and Fleet, David J and Norouzi, Mohammad and Salimans, Tim},
  journal={Journal of Machine Learning Research},
  volume={23},
  number={47},
  pages={1--33},
  year={2022}
}

@inproceedings{esser2023structure,
  title={Structure and content-guided video synthesis with diffusion models},
  author={Esser, Patrick and Chiu, Johnathan and Atighehchian, Parmida and Granskog, Jonathan and Germanidis, Anastasis},
  booktitle={Proceedings of the IEEE/CVF International Conference on Computer Vision},
  pages={7346--7356},
  year={2023}
}

@article{guo2023animatediff,
  title={AnimateDiff: Animate Your Personalized Text-to-Image Diffusion Models without Specific Tuning},
  author={Guo, Yuwei and Yang, Ceyuan and Rao, Anyi and Liang, Zhengyang and Wang, Yaohui and Qiao, Yu and Agrawala, Maneesh and Lin, Dahua and Dai, Bo},
  journal={International Conference on Learning Representations},
  year={2024}
}

@article{blattmann2023stable,
  title={Stable video diffusion: Scaling latent video diffusion models to large datasets},
  author={Blattmann, Andreas and Dockhorn, Tim and Kulal, Sumith and Mendelevitch, Daniel and Kilian, Maciej and Lorenz, Dominik and Levi, Yam and English, Zion and Voleti, Vikram and Letts, Adam and others},
  journal={arXiv preprint arXiv:2311.15127},
  year={2023}
}

@inproceedings{radford2021learning,
  title={Learning transferable visual models from natural language supervision},
  author={Radford, Alec and Kim, Jong Wook and Hallacy, Chris and Ramesh, Aditya and Goh, Gabriel and Agarwal, Sandhini and Sastry, Girish and Askell, Amanda and Mishkin, Pamela and Clark, Jack and others},
  booktitle={International conference on machine learning},
  pages={8748--8763},
  year={2021},
  organization={PMLR}
}

@inproceedings{zhang2023adding,
  title={Adding conditional control to text-to-image diffusion models},
  author={Zhang, Lvmin and Rao, Anyi and Agrawala, Maneesh},
  booktitle={Proceedings of the IEEE/CVF International Conference on Computer Vision},
  pages={3836--3847},
  year={2023}
}

@inproceedings{mou2024t2i,
  title={T2i-adapter: Learning adapters to dig out more controllable ability for text-to-image diffusion models},
  author={Mou, Chong and Wang, Xintao and Xie, Liangbin and Wu, Yanze and Zhang, Jian and Qi, Zhongang and Shan, Ying},
  booktitle={Proceedings of the AAAI Conference on Artificial Intelligence},
  volume={38},
  number={5},
  pages={4296--4304},
  year={2024}
}

@inproceedings{wang2024disco,
  title={Disco: Disentangled control for realistic human dance generation},
  author={Wang, Tan and Li, Linjie and Lin, Kevin and Zhai, Yuanhao and Lin, Chung-Ching and Yang, Zhengyuan and Zhang, Hanwang and Liu, Zicheng and Wang, Lijuan},
  booktitle={Proceedings of the IEEE/CVF Conference on Computer Vision and Pattern Recognition},
  pages={9326--9336},
  year={2024}
}

@article{song2020denoising,
  title={Denoising diffusion implicit models},
  author={Song, Jiaming and Meng, Chenlin and Ermon, Stefano},
  journal={arXiv preprint arXiv:2010.02502},
  year={2020}
}

@inproceedings{kingma2014auto,
  title={Auto-Encoding Variational Bayes},
  author={Kingma, Diederik P. and Welling, Max},
  booktitle={2nd International Conference on Learning Representations (ICLR)},
  year={2014}
}

@inproceedings{vaswani2017attention,
  author    = {Ashish Vaswani and Noam Shazeer and Niki Parmar and Jakob Uszkoreit and Llion Jones and Aidan N Gomez and Łukasz Kaiser and Illia Polosukhin},
  title     = {Attention is All You Need},
  booktitle = {Advances in Neural Information Processing Systems},
  volume    = {30},
  year      = {2017}
}

@inproceedings{lugaresi2019mediapipe,
  author    = {Camillo Lugaresi and Jiuqiang Tang and Hadon Nash and Chris McClanahan and Esha Uboweja and Michael Hays and Fan Zhang and Chuo-Ling Chang and Ming Yong and Juhyun Lee and Wan-Teh Chang and Wei Hua and Manfred Georg and Matthias Grundmann},
  title     = {MediaPipe: A Framework for Perceiving and Processing Reality},
  booktitle = {Proceedings of the Third Workshop on Computer Vision for AR/VR at IEEE Computer Vision and Pattern Recognition (CVPR)},
  year      = {2019}
}

@inproceedings{xu2020understanding,
  author = {Xu, Kun and Li, Chongxuan and Zhu, Jun and Zhang, Bo},
  title = {Understanding and Stabilizing GANs' Training Dynamics Using Control Theory},
  year = {2020},
  booktitle = {Proceedings of the 37th International Conference on Machine Learning (ICML)},
  publisher = {JMLR.org}
}

@article{wang2004image,
  title={Image quality assessment: from error visibility to structural similarity},
  author={Wang, Zhou and Bovik, Alan C and Sheikh, Hamid R and Simoncelli, Eero P},
  journal={IEEE Transactions on Image Processing},
  volume={13},
  number={4},
  pages={600--612},
  year={2004},
  publisher={IEEE}
}

@inproceedings{hore2010image,
  title={Image quality metrics: PSNR vs. SSIM},
  author={Hore, Alain and Ziou, Djemel},
  booktitle={2010 20th International Conference on Pattern Recognition},
  pages={2366--2369},
  year={2010},
  organization={IEEE}
}

@inproceedings{richardson2021encoding,
  title={Encoding in style: a stylegan encoder for image-to-image translation},
  author={Richardson, Elad and Alaluf, Yuval and Patashnik, Or and Nitzan, Yotam and Azar, Yaniv and Shapiro, Stav and Cohen-Or, Daniel},
  booktitle={Proceedings of the IEEE/CVF Conference on Computer Vision and Pattern Recognition},
  pages={2287--2296},
  year={2021}
}

@inproceedings{huang2020curricularface,
  title={Curricularface: adaptive curriculum learning loss for deep face recognition},
  author={Huang, Yuge and Wang, Yuhan and Tai, Ying and Liu, Xiaoming and Shen, Pengcheng and Li, Shaoxin and Li, Jilin and Huang, Feiyue},
  booktitle={proceedings of the IEEE/CVF conference on computer vision and pattern recognition},
  pages={5901--5910},
  year={2020}
}

@inproceedings{deng2019arcface,
  title={Arcface: Additive angular margin loss for deep face recognition},
  author={Deng, Jiankang and Guo, Jia and Xue, Niannan and Zafeiriou, Stefanos},
  booktitle={Proceedings of the IEEE/CVF conference on computer vision and pattern recognition},
  pages={4690--4699},
  year={2019}
}

@inproceedings{unterthiner2019towards,
  title={Towards accurate generative models of video: A new metric \& challenges},
  author={Unterthiner, Thomas and van Steenkiste, Sjoerd and Kurach, Karol and Marinier, Raphael and Michalski, Marcin and Gelly, Sylvain},
  booktitle={International Conference on Learning Representations (ICLR)},
  year={2019}
}

@inproceedings{kirschstein2024diffusionavatars,
  title={Diffusionavatars: Deferred diffusion for high-fidelity 3d head avatars},
  author={Kirschstein, Tobias and Giebenhain, Simon and Nie{\ss}ner, Matthias},
  booktitle={Proceedings of the IEEE/CVF Conference on Computer Vision and Pattern Recognition},
  pages={5481--5492},
  year={2024}
}

@inproceedings{qian2024gaussianavatars,
  title={Gaussianavatars: Photorealistic head avatars with rigged 3d gaussians},
  author={Qian, Shenhan and Kirschstein, Tobias and Schoneveld, Liam and Davoli, Davide and Giebenhain, Simon and Nie{\ss}ner, Matthias},
  booktitle={Proceedings of the IEEE/CVF Conference on Computer Vision and Pattern Recognition},
  pages={20299--20309},
  year={2024}
}

@article{chung2018voxceleb2,
  title={Voxceleb2: Deep speaker recognition},
  author={Chung, Joon Son and Nagrani, Arsha and Zisserman, Andrew},
  journal={arXiv preprint arXiv:1806.05622},
  year={2018}
}

@article{tu2024stableanimator,
  title={StableAnimator: High-Quality Identity-Preserving Human Image Animation},
  author={Tu, Shuyuan and Xing, Zhen and Han, Xintong and Cheng, Zhi-Qi and Dai, Qi and Luo, Chong and Wu, Zuxuan},
  journal={arXiv preprint arXiv:2411.17697},
  year={2024}
}

@inproceedings{yang2023effective,
  title={Effective whole-body pose estimation with two-stages distillation},
  author={Yang, Zhendong and Zeng, Ailing and Yuan, Chun and Li, Yu},
  booktitle={Proceedings of the IEEE/CVF International Conference on Computer Vision},
  pages={4210--4220},
  year={2023}
}

@inproceedings{jafarian2021learning,
  title={Learning high fidelity depths of dressed humans by watching social media dance videos},
  author={Jafarian, Yasamin and Park, Hyun Soo},
  booktitle={Proceedings of the IEEE/CVF Conference on Computer Vision and Pattern Recognition},
  pages={12753--12762},
  year={2021}
}

@inproceedings{balaji2019conditional,
  title={Conditional GAN with Discriminative Filter Generation for Text-to-Video Synthesis.},
  author={Balaji, Yogesh and Min, Martin Renqiang and Bai, Bing and Chellappa, Rama and Graf, Hans Peter},
  booktitle={IJCAI},
  volume={1},
  number={2019},
  pages={2},
  year={2019}
}

@InProceedings{xie2022vfhq,
      author = {Liangbin Xie and Xintao Wang and Honglun Zhang and Chao Dong and Ying Shan},
      title = {VFHQ: A High-Quality Dataset and Benchmark for Video Face Super-Resolution},
      booktitle={The IEEE Conference on Computer Vision and Pattern Recognition Workshops (CVPRW)},
      year = {2022}
  }

@article{ephrat2018looking,
  title={Looking to listen at the cocktail party: a speaker-independent audio-visual model for speech separation},
  author={Ephrat, Ariel and Mosseri, Inbar and Lang, Oran and Dekel, Tali and Wilson, Kevin and Hassidim, Avinatan and Freeman, William T. and Rubinstein, Michael},
  journal={ACM Trans. Graph.},
  volume={37},
  number={4},
  articleno={112},
  year={2018}
}

@inproceedings{zhang2021flow,
  title={Flow-Guided One-Shot Talking Face Generation With a High-Resolution Audio-Visual Dataset},
  author={Zhang, Zhimeng and Li, Lincheng and Ding, Yu and Fan, Changjie},
  booktitle={Proceedings of the IEEE/CVF Conference on Computer Vision and Pattern Recognition},
  pages={3661--3670},
  year={2021}
}

@inproceedings{yi2023generating,
title={Generating Holistic 3D Human Motion from Speech},
  author={Yi, Hongwei and Liang, Hualin and Liu, Yifei and Cao, Qiong and Wen, Yandong 
and Bolkart, Timo and Tao, Dacheng and Black, Michael J},
booktitle={IEEE Conference on Computer Vision and Pattern Recognition (CVPR)}, 
pages={469-480},
month={June}, 
year={2023} 
}

@article{yin2023nuwa,
  title={Nuwa-xl: Diffusion over diffusion for extremely long video generation},
  author={Yin, Shengming and Wu, Chenfei and Yang, Huan and Wang, Jianfeng and Wang, Xiaodong and Ni, Minheng and Yang, Zhengyuan and Li, Linjie and Liu, Shuguang and Yang, Fan and others},
  journal={arXiv preprint arXiv:2303.12346},
  year={2023}
}

@article{wang2024magicvideo,
  title={Magicvideo-v2: Multi-stage high-aesthetic video generation},
  author={Wang, Weimin and Liu, Jiawei and Lin, Zhijie and Yan, Jiangqiao and Chen, Shuo and Low, Chetwin and Hoang, Tuyen and Wu, Jie and Liew, Jun Hao and Yan, Hanshu and others},
  journal={arXiv preprint arXiv:2401.04468},
  year={2024}
}

@inproceedings{salimans2022progressive,
  title={Progressive Distillation for Fast Sampling of Diffusion Models},
  author={Salimans, Tim and Ho, Jonathan},
  booktitle={International Conference on Learning Representations (ICLR)},
  year={2022},
}

@article{wang2024yolov10,
  title={Yolov10: Real-time end-to-end object detection},
  author={Wang, Ao and Chen, Hui and Liu, Lihao and Chen, Kai and Lin, Zijia and Han, Jungong and others},
  journal={Advances in Neural Information Processing Systems},
  volume={37},
  pages={107984--108011},
  year={2024}
}

@article{deng2019retinaface,
  title={Retinaface: Single-stage dense face localisation in the wild},
  author={Deng, Jiankang and Guo, Jia and Zhou, Yuxiang and Yu, Jinke and Kotsia, Irene and Zafeiriou, Stefanos},
  journal={arXiv preprint arXiv:1905.00641},
  year={2019}
}

@inproceedings{men2025mimo,
  title={Mimo: Controllable character video synthesis with spatial decomposed modeling},
  author={Men, Yifang and Yao, Yuan and Cui, Miaomiao and Bo, Liefeng},
  booktitle={Proceedings of the Computer Vision and Pattern Recognition Conference},
  pages={21181--21191},
  year={2025}
}

@article{hu2025animateanyone2,
      title={Animate Anyone 2: High-Fidelity Character Image Animation with Environment Affordance},
      author={Li Hu and Guangyuan Wang and Zhen Shen and Xin Gao and Dechao Meng and Lian Zhuo and Peng Zhang and Bang Zhang and Liefeng Bo},
      journal={arXiv preprint arXiv:2502.06145},
      year={2025}
}

@inproceedings{chung2016out,
  title={Out of time: automated lip sync in the wild},
  author={Chung, Joon Son and Zisserman, Andrew},
  booktitle={Asian conference on computer vision},
  pages={251--263},
  year={2016},
  organization={Springer}
}

@inproceedings{liu2025edcflow,
  title={EDCFlow: Exploring Temporally Dense Difference Maps for Event-based Optical Flow Estimation},
  author={Liu, Daikun and Cheng, Lei and Wang, Teng and Sun, Changyin},
  booktitle={Proceedings of the Computer Vision and Pattern Recognition Conference},
  pages={1984--1993},
  year={2025}
}
